# A Pose-only Solution to Visual Reconstruction and Navigation

Qi Cai[1]†, Lilian Zhang[2]†, Yuanxin Wu[1]†*, Wenxian Yu[1], Dewen Hu[2]*

## Abstract

*Visual navigation and three-dimensional (3D) scene reconstruction are essential for robotics to interact with the surrounding environment. Large-scale scenes and critical camera motions are great challenges facing the research community to achieve this goal. We raised a pose-only imaging geometry framework and algorithms that can help solve these challenges. The representation is a linear function of camera global translations, which allows for efficient and robust camera motion estimation. As a result, the spatial feature coordinates can be analytically reconstructed and do not require nonlinear optimization. Experiments demonstrate that the computational efficiency of recovering the scene and associated camera poses is significantly improved by 2-4 orders of magnitude. This solution might be promising to unlock real-time 3D visual computing in many forefront applications.*

## 1. Introduction

A visual imaging system maps the 3D real world onto a two-dimensional image camera plane. One essential task of computer vision research is to recover a 3D scene and the camera poses at which the images were taken. As noted by Marr [1], humans perceive the real world through two main processes: image feature correspondence, followed by the computation and understanding of the 3D scene. The reverse-imaging process of recovering the scene and the associated camera poses from a set of images, known as bundle adjustment (BA), is the backbone of simultaneous localization and mapping or structure from motion, and it plays a prominent role in computer vision, robotics, and digital photogrammetry applications [2-6]. BA is essentially an iterative nonlinear optimization with respect to 3D feature coordinates and camera poses (sometimes including intrinsic camera parameters) [2, 7]; its performance heavily depends on initialization [8-13]. However, special but not uncommon camera movements, such as collinear or small translations, typically lead to abnormal initialization [7, 12, 14, 15].

Visual computation efficiency and robustness have been long-standing bottleneck problems in 3D computer vision. Specifically, the nonlinear optimization of a large-scale BA has been facing two challenges [3, 10, 11, 16]: benign initialization and fast solution to the normal equation. For initialization, an incremental optimization starting from a two-view BA can be employed [5, 17], or alternatively, the relative poses between any two views can be used as inputs for optimally solving first the global rotation and then the global translation of each view. There exist several efficient and stable global rotation averaging methods [13, 18-23].

The relative translations can be formulated as $\lambda t_{i,j} = R_j (t_i - t_j)$ where $\lambda = \|t_i - t_j\|$, $t_{i,j}$ is a relative translation unit vector between the i-th and j-th views, and $R_i$ and $t_i$ are the global rotation and translation for the i-th view, respectively. A direct linear approach by Govindu [20] proposed a least-squared solution of global translation to the linear system $t_{i,j} \times R_j (t_i - t_j) = 0$ and improved it in [24]. Recent global translation averaging methods in [12, 15, 25-28] aimed to minimize the penalty of $\lambda t_{i,j} - R_j (t_i - t_j)$ or its variants. Sim and Hartley [28] and Moulon et al. [27] formulated the penalty as a $L_\infty$ norm. Wilson and Snavely [12] introduced a 1DSfM method, which has a one-dimensional preprocessing step to remove relative translation outliers and uses a non-convex optimization in the squared chordal form. Ozyesil [29] pointed out that the $L_\infty$ norm is prone to pairwise translation outliers, and provided a robust penalty with a least unsquared deviations (LUD) form [15]. Goldstein et al. [26] described the penalty by the magnitude of the projection of $t_i - t_j$ on the orthogonal complement of $t_{i,j}$ and minimized it by the alternating direction method of multipliers. Zhuang et al. [25] gave a geometric interpretation of the above-mentioned works and developed a bilinear angle-based translation averaging (BATA) method. These methods might suffer from collinear motion, parallel rigidity [15], and even the local pure rotation motion. Other recent heuristic methods in global translation averaging construct different objective functions from $\lambda t_{i,j} - R_j (t_i - t_j)$. In [14], a global linear method presents a coplanar constraint on triple views. Cui [30] added directions of features to the above constraint to help deal with collinear motion. The recent work by Liu et al. [31] showed that the idea of using feature directions can well handle the collinear motion, and presented a linear translation form by calculating a non-linear term directly based on $t_{i,j}$. Consequently, the accuracy of current global translation averaging methods in all the above-mentioned works depends on that of relative translation.

[1] 1 Shanghai Key Laboratory of Navigation and Location-based Services, School of Electronic Information and Electrical Engineering, Shanghai Jiao Tong University, Shanghai, China.
2 College of Intelligent Science and Technology, National University of Defense Technology, Changsha, China.
† These authors contributed equally to this work.
* Correspondence to: yuanxin.wu@sjtu.edu.cn, dwhu@nudt.edu.cn.

For a fast solution to the large-scale normal equation, the main ideas for the last two decades have been full utilization of the inherent sparsity property of the BA problem, and the reduction of matrix dimension by the Schur complement [3, 11, 32, 33]. There are works reformulating the BA problem by using different parameterization, such as the parallax angles [11, 31, 34], to express 3D feature coordinates. However, the high-dimensional parameter space in the BA problem still exists. To overcome the memory limit of a computer, the BA problem was transformed into a number of small-scale inter-connected BA problems and handled by distributed computers [8, 10].

Based on the given feature correspondence, Higgins [35] introduced the concept of essential matrix and invented a linear eight-point algorithm to recover the two-view pose and scene structure. The essential equation defining the essential matrix is a simplification of the two-view imaging geometry in that it only captures the co-planar relationship of the camera baseline and the two projection rays, but loses the depth information [2, 7, 35-37]. It was recently revealed that the two-view imaging geometry is equivalently governed by a pair of pose-only constraints, decoupling camera poses from 3D feature coordinates [36].

In this paper, we find that the multiple-view imaging geometry can be completely represented by camera poses and image points, and notably, it is linearly related to camera global translations (see Fig. 1). Preconditioned on known global rotations, we give a linear global translation solution without the need of relative translation that can deal with the motions of collinear and local pure rotation. This linear translation relationship is found to be instrumental in obtaining a nearly optimal initialization for the subsequent nonlinear optimization. Over 50 data tests on public datasets show that the proposed algorithms have considerably eased the challenges of computational efficiency and robustness in recovering camera poses and the 3D scene structure.

## 2. Pose-only imaging geometry

### 2.1. Depth-pose-only constraint

Consider a 3D feature point $X^W = (x^W, y^W, z^W)^T$ observed in $n$ images (or views). For $i = 1, 2, …, n$, denote by $X_i = (x_i, y_i, 1)^T$ the normalized image coordinate of the 3D feature point in the $i$-th image (or, alternatively, view $i$), and by $R_i$ and $t_i$ the global rotation and global translation of the camera when taking the $i$-th image, respectively. The projection equation of the 3D feature point $X^W$ for the $i$-th image can be given by [2, 7]

$$X_i = \frac{1}{z^{C_i}} X^{C_i} = \frac{1}{z^{C_i}} R_i (X^W - t_i), \quad i = 1, 2, ..., n \quad (1)$$

where $X^{C_i} = (x^{C_i}, y^{C_i}, z^{C_i})^T$ is the coordinate of the 3D feature point in the camera frame corresponding to the $i$-th image, and $z^{C_i} > 0$ is the corresponding depth of the feature point. For $m$ 3D feature points observed in $n$ images, the multiple-view imaging relationship can be represented as [2, 7]

$$f\left(\{X_k^w\}_{k=1...m}, \{R_i, t_i\}_{i=1...n}, \{X_{k,i}\}_{k=1...m, i=1...n}\right) = 0 \quad (2)$$

where $X_k^w$ is the world coordinate of the $k$-th 3D feature; $R_i$ and $t_i$ denote the global rotation and translation of the camera when taking the $i$-th image, respectively; and $X_{k,i}$ is the normalized image coordinate of the $k$-th 3D feature on the $i$-th image. It is well known that there is a global scale ambiguity in recovering camera poses and the 3D scene structure. For instance, for any rigid transformation $R$ and $t$ at a scale $\alpha$, the projection equation (1) is always valid for substitutions $R_i \to R_i R^T$, $t_i \to \alpha R(t_i + t)$, and $X^W \to \alpha R(X^W + t)$. Therefore, the discussions to follow are based on global scale ambiguity awareness. Denote by $(i, j)$ a pair of views consisting of the $i$-th and $j$-th images. The imaging equation for the view pair $(i, j)$ is [2, 7]

$$z^{C_j} \mathbf{X}_j = z^{C_i} R_{i,j} \mathbf{X}_i + \mathbf{t}_{i,j} \quad (3)$$

where the relative rotation is $R_{i,j} = R_j R_i^T$ and the relative translation is $\mathbf{t}_{i,j} = R_j(\mathbf{t}_i - \mathbf{t}_j)$. Left multiply the antisymmetric matrix $[\mathbf{X}_j]_\times$ on both sides of equation (3),

$$z^{C_i} [\mathbf{X}_j]_\times R_{i,j} \mathbf{X}_i = -[\mathbf{X}_j]_\times \mathbf{t}_{i,j} \quad (4)$$

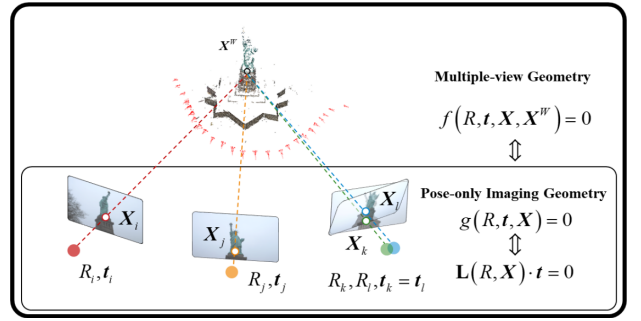

Figure 1: Principle of the proposed solution. Multiple-view geometry is equivalently represented by poses and image points, which is actually a linear global translation constraint. The linear constraint enables a linear solution to global translations that is theoretically immune to camera collinear movement and local pure rotation. 3D feature coordinates are removed from optimization in pose-only imaging geometry and can be analytically reconstructed from recovered poses.

Taking the magnitude, we get

$$z^{C_i} = \frac{\|[\mathbf{X}_j]_\times \mathbf{t}_{i,j}\|}{\theta_{i,j}} \triangleq d_i^{(i,j)} \quad (5)$$

where $\theta_{i,j} = \|[\mathbf{X}_j]_\times R_{i,j} \mathbf{X}_i\|$. Similarly, left-multiplying the antisymmetric matrix $[R_{i,j}\mathbf{X}_i]_\times$ on both sides of equation (3) yields

$$z^{C_j} = \frac{\|[R_{i,j}\mathbf{X}_i]_\times \mathbf{t}_{i,j}\|}{\theta_{i,j}} \triangleq d_j^{(i,j)} \quad (6)$$

Combining equations (3), (5), and (6), the pose-only constraint for the two-view imaging geometry, called a pair of pose-only or PPO constraints [36], is obtained as

$$d_j^{(i,j)} \mathbf{X}_j = d_i^{(i,j)} R_{i,j} \mathbf{X}_i + \mathbf{t}_{i,j} \quad (7)$$

Moreover, it can be proved that the PPO constraint is equivalent to the two-view imaging geometry [36]. This equivalency is valid even when there is only a pure rotation between the two views, namely, in the case of $\theta_{i,j} = 0$. Regarding the $l$-th image ($l \neq j$), the view pair $(i,l)$ also satisfies the PPO constraint

$$d_l^{(i,l)} \mathbf{X}_l = d_i^{(i,l)} R_{i,l} \mathbf{X}_i + \mathbf{t}_{i,l} \quad (8)$$

and

$$d_i^{(i,j)} = d_i^{(i,l)} = z^{C_i} \quad (9)$$

We name the relationship in equation (9) as the *depth-equal constraint* of the 3D feature point on the $i$-th image. Note that for all $n$ images, there are $C_n^2$ PPO constraints and $C_n^3$ depth-equal constraints, which contain a great deal of redundancy.

Substitute equation (9) into equation (8),

$$d_l^{(i,l)} \mathbf{X}_l = d_i^{(i,j)} R_{i,l} \mathbf{X}_i + \mathbf{t}_{i,l} \quad (10)$$

Define a set

$$D(\varsigma, \eta) = \left\{ d_i^{(\varsigma,i)} \mathbf{X}_i = d_\varsigma^{(\varsigma,\eta)} R_{\varsigma,i} \mathbf{X}_\varsigma + \mathbf{t}_{\varsigma,i} \mid 1 \leq i \leq n, i \neq \varsigma \right\} \quad (11)$$

which represents a set of constraints that take views $\varsigma$ and $\eta$ as the *left-* and *right-base views*, respectively. As this is related to poses and depths (which are functions of poses) only, we name it the depth-pose-only (DPO) constraint set for the 3D feature point. Note: It can be proved that the DPO constraint set (11) is equivalent to the projection equation (1); see Proposition 3 below. That is, For $m$ 3D feature points observed in $n$ images, the multiple-view imaging relationship (2) can be equivalently expressed in a pose-only form

$$g\left(\{R_i, \mathbf{t}_i\}_{i=1...n}, \{\mathbf{X}_{k,i}\}_{k=1...m, i=1...n}\right) = 0 \quad (12)$$

According to Proposition 2 below, the two-view PPO constraint (7) can be rewritten as a linear form of relative translation

$$\left(\mathbf{X}_j \mathbf{b}_{i,j}^T - R_{i,j} \mathbf{X}_i \mathbf{a}_{i,j}^T - \theta_{i,j}^2 I_3\right) \mathbf{t}_{i,j} = 0 \quad (13)$$

By analogy, the multiple-view DPO constraint (11) can also be linearly expressed in terms of relative translation

$$\theta_{\varsigma,i}^2 R_{\varsigma,i} \mathbf{X}_\varsigma \mathbf{a}_{\varsigma,\eta}^T \mathbf{t}_{\varsigma,\eta} + \theta_{\varsigma,i}^2 \theta_{\varsigma,\eta}^2 \mathbf{t}_{\varsigma,i} - \theta_{\varsigma,\eta}^2 \mathbf{X}_i \mathbf{b}_{\varsigma,i}^T \mathbf{t}_{\varsigma,i} = 0 \quad (14)$$

Alternatively, the above expressions can be readily expressed in terms of global translation. In the sequel, however, we will present another linear expression of the global translation.

## 2.2. Linear global translation constraint

Currently, global rotation initialization algorithms, such as the rotation averaging algorithm proposed by Chatterjee and Govindu [13], perform fairly well. The remaining sub-section attempts to solve global translations preconditioned on known global rotations $\{R_i\}_{i=1...n}$.

Left multiply $[\mathbf{X}_i]_\times$ on both sides of the DPO constraint (11),

$$0 = [\mathbf{X}_i]_\times \left(d_\varsigma^{(\varsigma,\eta)} R_{\varsigma,i} \mathbf{X}_\varsigma + \mathbf{t}_{\varsigma,i}\right), \quad 1 \leq i \leq n, i \neq \varsigma \quad (15)$$

According to Proposition 2 below,

$$d_\varsigma^{(\varsigma,\eta)} = \frac{\mathbf{a}_{\varsigma,\eta}^T \mathbf{t}_{\varsigma,\eta}}{\theta_{\varsigma,\eta}^2} \quad (16)$$

Substituting equation (16) into equation (15), we show that global translations satisfy the following linear homogeneous equation

$$B\mathbf{t}_\eta + C\mathbf{t}_i + D\mathbf{t}_\varsigma = 0, \quad 1 \leq i \leq n, i \neq \varsigma \quad (17)$$

in which

$$\begin{aligned} B &= [\mathbf{X}_i]_\times R_{\varsigma,i} \mathbf{X}_\varsigma \mathbf{a}_{\varsigma,\eta}^T R_\eta \\ C &= \theta_{\varsigma,\eta}^2 [\mathbf{X}_i]_\times R_i \\ D &= -(B+C) \end{aligned} \quad (18)$$

If $[\mathbf{X}_i]_\times R_{\varsigma,i} \mathbf{X}_\varsigma = 0$ for $1 \leq i \leq n, i \neq \varsigma$, then $B = C = D = 0$ and equation (17) is always true for any global translation.

For all 3D feature points, denote by $\mathbf{t} = (\mathbf{t}_1^T, ..., \mathbf{t}_n^T)^T$ the concatenated global translation of $n$ images and rewrite equation (17) as

$$\mathbf{L} \cdot \mathbf{t} = 0 \quad (19)$$

where **L** is a matrix comprising global rotations and normalized image coordinates. It can be proved that $rank(\mathbf{L}) = 3n-4$ when there are at least two 3D feature points satisfying $\theta_{\varsigma,\eta} \neq 0$ (see Proposition 6 below). Equation (19) is called the linear global translation (LiGT) constraint. Choose view $r$ as the global translation reference, namely,

$$\boldsymbol{t}_r = 0 \tag{20}$$

An estimate of global translation $\hat{\boldsymbol{t}}$ ($\boldsymbol{t}_r$ removed) can be obtained by solving the linear homogeneous equation (19).

There are two $\hat{\boldsymbol{t}}$ with opposite signs; however, the right one can be readily identified by using equation (16), that is, it should satisfy $\boldsymbol{a}_{\varsigma,\eta}^T \boldsymbol{t}_{\varsigma,\eta} \geq 0$.

Consequently, according to Propositions 3-5 below, the three representations of multiple-view imaging relationship, namely (2), (12), and (19), are equivalent. Equation (19) expresses the multiple-view imaging relationship as a linear constraint. Given global rotations, the LiGT constraint (19) enables a linear solution to global translations, which is proved to be theoretically immune to camera collinear movement and local pure rotation. Certainly, the accuracy of the obtained global translations would be affected by the quality of the given global rotations. If a higher pose accuracy is required, a proposed algorithm of pose adjustment (given in Section 3) can be used to further refine the camera poses. The 3D feature coordinates can be analytically recovered from the camera poses.

## 2.3. Propositions

*Proposition 1.* $d_i^{(i,j)} = d_i^{(j,i)}$ for view pair $(i, j)$

*Proposition 2.* The depth can be linearly expressed in terms of translation, that is, $d_i^{(i,j)} = \dfrac{\boldsymbol{a}_{i,j}^T \boldsymbol{t}_{i,j}}{\theta_{i,j}^2}$ and $d_j^{(i,j)} = \dfrac{\boldsymbol{b}_{i,j}^T \boldsymbol{t}_{i,j}}{\theta_{i,j}^2}$, where $\boldsymbol{a}_{i,j}^T = \left(\left[R_{i,j} X_i\right]_\times X_j\right)^T \left[X_j\right]_\times$ and $\boldsymbol{b}_{i,j}^T = \left(\left[R_{i,j} X_i\right]_\times X_j\right)^T \left[R_{i,j} X_i\right]_\times$.

*Proposition 3.* The DPO constraint set (11) $\Leftrightarrow$ the projection equation (1).

*Proposition 4.* The LiGT constraint $\Rightarrow$ the depth-equal constraint.

*Proposition 5.* The LiGT constraint $\Leftrightarrow$ the DPO constraint.

*Proposition 6.* When there are at least two 3D feature points with different image points such that $\theta_{\varsigma,\eta} = \|X_\eta \times R_{\varsigma,\eta} X_\varsigma\| \neq 0$, $rank(\mathbf{L}) = 3n-4$.

For a global pure rotation, $\theta_{\varsigma,\eta} = 0$ for all 3D feature points and the LiGT constraint can never generate the right global translation; neither can the projection equation, nor the DPO constraint. However, as a scenario violating the Proposition 6 precondition only occurs theoretically, Proposition 6 actually indicates that the global translation can almost certainly be solved from the LiGT constraint, even under special but common movements such as collinear motion or local pure rotation.

## 3. Pose adjustment

The gold-standard BA minimizes the reprojection error formulated by the projection equation (1). Denoting by $\tilde{X}_i$ the error-contaminated normalized image coordinate of a 3D feature point in the $i$-th image, the reprojection error is usually defined as

$$V_i = X_i - \tilde{X}_i = \frac{Y_i^{BA}}{\boldsymbol{e}_3^T Y_i^{BA}} - \tilde{X}_i \tag{21}$$

where $Y_i^{BA} = R_i(X^W - \boldsymbol{t}_i)$ and $\boldsymbol{e}_3^T = (0,0,1)$. For $m$ 3D feature points observed in $n$ images, a reprojection error vector $V_{BA}$ can be formed. The error function in the BA minimization can be expressed as

$$\varepsilon_{BA}\left(\{X_k^W\}_{k=1\ldots m}, \{R_i, \boldsymbol{t}_i\}_{i=1\ldots n}, \{\tilde{X}_{k,i}\}_{k=1\ldots m, i=1\ldots n}\right) = V_{BA}^T V_{BA} \tag{22}$$

The corresponding BA minimization problem is formulated as [2, 7]

$$\underset{\{X_k^W\}_{k=1\ldots m}, \{R_i, \boldsymbol{t}_i\}_{i=1\ldots n}}{\arg\min} \varepsilon_{BA} \tag{23}$$

As there are typically a large number of 3D features in a scene, we can imagine that the equation (23) is a nonlinear optimization problem in a high-dimensional parameter space. With the DPO constraint set in equation (11), the reprojection error for a 3D feature point is given by

$$V_i = X_i - \tilde{X}_i = \frac{Y_i^{PA}}{\boldsymbol{e}_3^T Y_i^{PA}} - \tilde{X}_i \tag{24}$$

where $Y_i^{PA} = \tilde{d}_\varsigma^{(\varsigma,\eta)} R_{\varsigma,i} \tilde{X}_\varsigma + \boldsymbol{t}_{\varsigma,i}$ and $\tilde{d}_\varsigma^{(\varsigma,\eta)}$ is computed using the error-contaminated normalized image coordinate. For $m$ 3D feature points observed in $n$ images, a reprojection error vector $V_{PA}$ can be formed. The error function in the minimization can be expressed as

$$\varepsilon_{PA}\left(\{R_i, \boldsymbol{t}_i\}_{i=1\ldots n}, \{\tilde{X}_{k,i}\}_{k=1\ldots m, i=1\ldots n}\right) = V_{PA}^T V_{PA} \tag{25}$$

The corresponding minimization problem is formulated as

$$\underset{\{R_i, \boldsymbol{t}_i\}_{i=1\ldots n}}{\arg\min} \varepsilon_{PA} \tag{26}$$

the unknown parameters of which consist of camera poses only. Therefore, it is referred to as pose adjustment (PA) throughout the paper.

### 3.1. Global analytical reconstruction

The 3D multiple-view scene structure can be analytically reconstructed from the obtained camera poses. For a 3D feature point, its depth in the left-base view is calculated as

$$\hat{z}_\varsigma^W = \sum_{\substack{1 \le i \le n \\ i \ne \varsigma}} \omega_{\varsigma,i} d_\varsigma^{(\varsigma,i)} \quad (27)$$

where $\omega_{\varsigma,i}$ is the weighting coefficient. According to the two-view case [36], $\theta_{\varsigma,i}$ is a quality indicator of reconstruction, and thus, we take the weighting coefficient as $\omega_{\varsigma,i} = \theta_{\varsigma,i} / \sum_{\substack{1 \le i \le n \\ i \ne \varsigma}} \theta_{\varsigma,i}$. Finally, the 3D feature coordinate is given by

$$X^W = \hat{z}_\varsigma^W R_\varsigma^T X_\varsigma + t_\varsigma \quad (28)$$

The world coordinates of $m$ 3D features observed in $n$ images can be represented entirely by camera poses and image points as $\{X_k^w\}_{k=1...m} = z\left(\{R_i, t_i\}_{i=1...n}, \{X_{k,i}\}_{k=1...m, i=1...n}\right)$.

### 3.2. Pose-only algorithm for recovering camera poses and the 3D scene

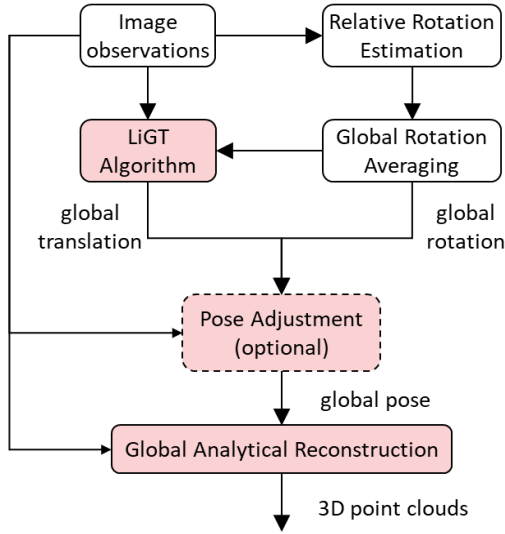

Figure 2: Flow chart of pose-only algorithm

For $m$ 3D feature points and $N$ images (or views). Note that all feature points are not necessarily observed in each image here.

*Inputs*: Global rotations and normalized image coordinates.

*Step 1.* Designate a view, say view $r$, as the reference view. Set the constraint $t_r = 0$.

*Step 2.* For the current 3D feature point $X^W$, select left/right-base views using the following criterion

$$(\varsigma, \eta) = \arg\max_{1 \le i,j \le n} \{\theta_{i,j}\} \quad (29)$$

*Step 3.* Build the matrix $L$ using equations (17) and (18).

*Step 4.* For all 3D feature points, repeat Steps 2-3. Obtain the global translation $\hat{t}$ by solving equation (19).

*Step 5.* Identify the right global translation solution using $a_{\varsigma,\eta}^T t_{\varsigma,\eta} \ge 0$.

*Step 6* (Optional). Implement PA to further improve camera poses according to equation (26).

*Step 7.* Analytically reconstruct all 3D feature coordinates using equations (27) and (28).

The flow chart of the pose-only algorithm is given in Fig. 2.

## 4. Experiments

The experiment was performed on an Ubuntu 18.04.4 LTS platform, with 128 GB memory and Intel® Xeon(R) Platinum 8269CY CPU @ 2.50 GHz, one core. The LiGT algorithm was developed based on Spectra and Eigen C++ libraries, and the PA algorithm was developed using the SparseLM optimization library.

*Two-view relative pose.* We utilized the OpenGV library, which comprises various common two-view processing algorithms [35, 38-41]

*Global rotation.* The state-of-the-art libraries of global structure from motion (SfM), such as OpenMVG and Theia Vision [42], are mainly based on algorithms proposed by Chatterjee [19], Hartley [21, 22], and Martinec [43]. Comparably, Chatterjee's newest algorithm [13] has the best accuracy and robustness [12, 15] and thus was used to provide the global rotation of each view.

*Global translation.* Recovering the global translation is a key problem in SfM. We compared LUD [15], 1DSfM [12], and linearSfM [14] in the Theia Vision library. LUD is found to be robust and outstanding, and thus, it is mainly shown in real data tests.

*Global optimization.* BA is the cornerstone of visual geometry computation. Most state-of-the-art libraries incorporate the Google Ceres BA [44], although there have been a number of studies focused on speeding-up BA, such as sBA [3], ssBA [45], and PBA (or PMBA) [11, 31]. This study addresses the standard case of the calibrated camera; therefore, the Google Ceres BA (version 1.14.0) was taken as the benchmark for algorithm assessment.

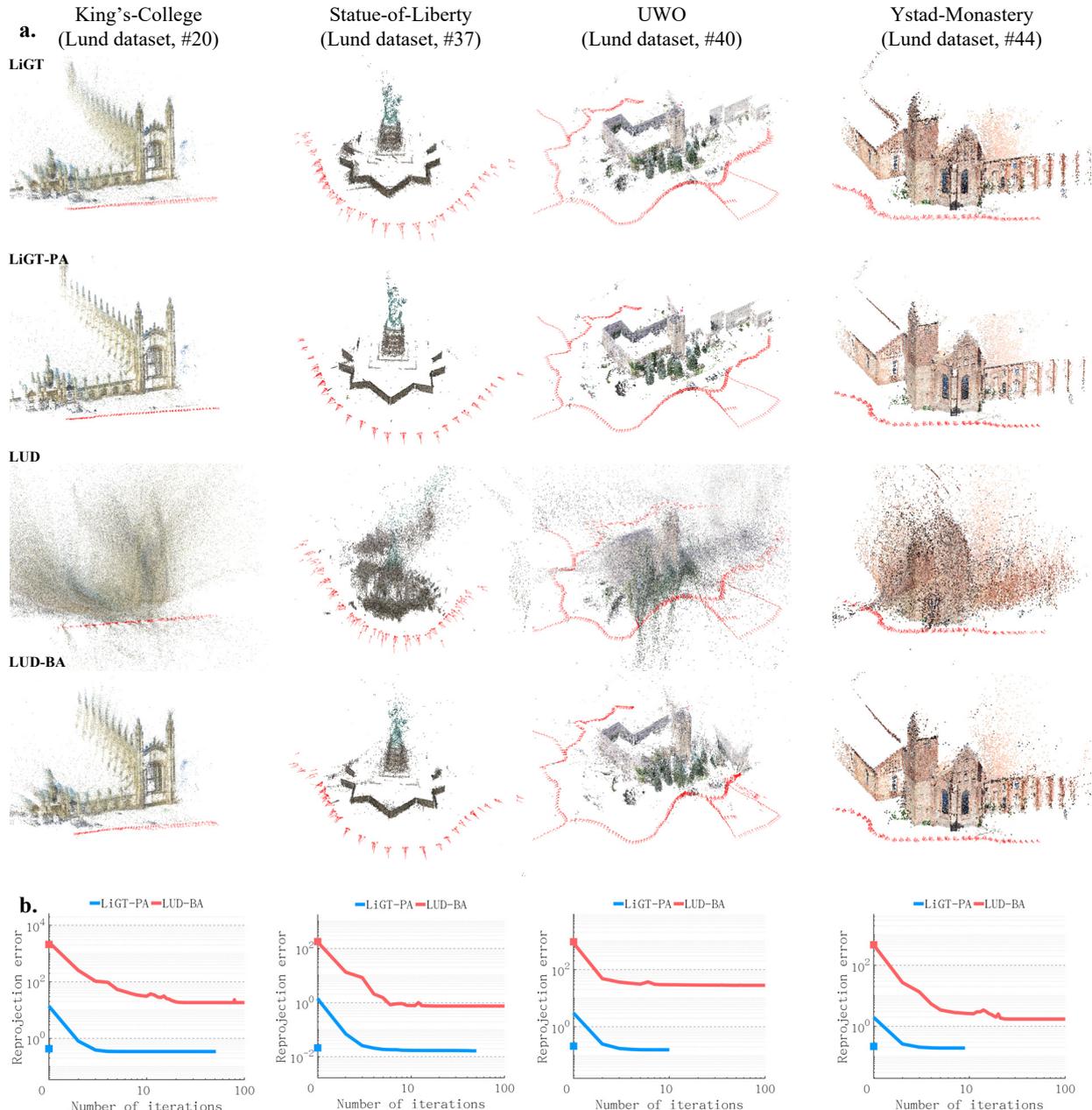

Figure 3: Recovered camera poses and 3D scenes, and reprojection errors of representative data. King's-College, Statue-of-Liberty, UWO and Ystad-Monastery are from the Lund dataset. **a**, Recovered camera poses and 3D scenes by LiGT, LiGT-PA, LUD, and LUD-BA. Red arrows denote cameras; **b**, reprojection errors for LiGT-PA and LUD-BA as a function of the number of iterations (maximum set at 100) performed during the optimization process. The 3D scenes were recovered analytically in LiGT and LiGT-PA, and by traditional triangulation in LUD. The squares on each vertical axis denote the reprojection errors of LUD and LiGT when global rotations refined by LiGT-PA are used as the input instead.

## 4.1. Real test performance

We performed a number of tests on 55 data from the Lund and OpenSLAM public datasets [46, 47].

Figure 3 presents four representative test results of King's College Cambridge, Statue of Liberty, University of Western Ontario, and Monastery in Ystad from the Lund dataset. The rotation averaging algorithm [13] has the best accuracy and robustness [12, 15] and, thus, has been used to provide global rotations. The LiGT algorithm, as well as the PA algorithm initialized with the LiGT algorithm (LiGT-PA), are compared against state-of-the-art counterparts: the LUD algorithm [15], to determine global translations, and the Google Ceres BA algorithm [44] initialized with the LUD algorithm (LUD-BA). The LiGT reprojection error is significantly smaller, by approximately two orders of magnitude, than that of LUD. The results show that the LiGT 3D scene recovery is very close to that of LUD-BA or LiGT-PA, in contrast to the LUD result that barely shows the scene outlines. The LiGT-PA algorithm, within fewer iterations, leads to a reprojection error that is 1-2 orders of magnitude smaller than that of LUD-BA. The scene recovery of LUD-BA appears to be incomplete (e.g., the Statue of Liberty goddess body and the University of Western Ontario wall and camera poses). In Fig. 3b, the two squares lying on each vertical axis denote the reprojection errors of LUD and LiGT when global rotations refined by LiGT-PA are used as the input instead. The reprojection errors of LiGT are further reduced by 1-2 orders of magnitude, approaching those of LiGT-PA.

The computational costs (in terms of running time and memory cost) and the reprojection errors across all 55 data tests are summarized in Fig. 4 clockwise in ascending order of the number of image points. Compared with LUD-BA, LiGT-PA reduces, on average, the running time by approximately 25 times and the memory consumption by approximately 15 times, whereas LiGT significantly reduces the running time by approximately 8,000 times on average (140,000 times maximum) and the memory consumption by approximately 400 times on average (5,000 times maximum). It can be well predicted from the clockwise increasing trend in Fig. 4 that the LiGT's computational efficiency advantage will be significantly more prominent for larger-scale data. Furthermore, in terms of the final reprojection error, LiGT-PA consistently outperforms LUD-BA, and notably, the error of LiGT is even smaller than that of LUD-BA in several tests.

## 4.2. Further discussions

In summary, the 3D scene quality of the LiGT algorithm is very close to that of BA and PA. The reprojection error profile of all algorithms shows that the LiGT algorithm enables faster convergence and smaller reprojection errors for optimization in only a few iterations, compared with the LUD algorithm. Note that the reprojection errors have been regularized uniformly for all algorithms by way of BA's minimization function, using their own estimates of camera poses and 3D feature coordinates. In fact, it can be well predicted, according to Propositions 3 and 5, that if the provided global rotations have high accuracy, the global translation solution by the LiGT algorithm would be close to optimality.

*LiGT algorithm.* It appears that, under such special motions as small translation or collinear movement, the LiGT algorithm significantly outperforms the LUD algorithm in terms of both reprojection error and 3D scene quality. Notably, the LiGT's 3D scene can compete in appearance with those of BA and PA; for instance, in the case of closed-loop camera motions (#13: Eglise-interior and #38: The-Pumpkin). For more complex camera motions (#24: Linkoping-Cathedral), the projection error of the LiGT algorithm is even smaller than that of LUD-BA, the BA algorithm initialized by the LUD algorithm.

*PA algorithm.* LiGT-PA generally outperforms LUD-BA in terms of both reprojection errors and 3D scene quality. The LiGT-PA typically requires only a few iterations to reach convergence, and its final reprojection errors are smaller than those of LUD-BA (#8: Buddha-temple, #24: Linkoping-Cathedral, and #38: The-Pumpkin).

*Collinear motion.* Under collinear motions (#8: Buddha-temple, #9: De-Guerre, and #36: Sri-Veeramakaliamman) or local linear motion (#50:

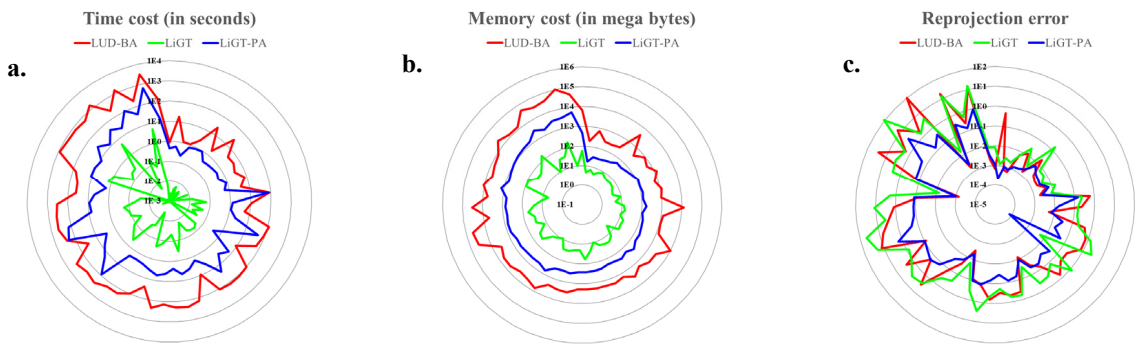

Figure 4: Running time, memory consumption, and final reprojection errors for LiGT, LUD-BA, and LiGT-PA. Arranged clockwise in ascending order of the number of image points for all 55 data results. **a,** Time cost in seconds; **b,** Memory cost in megabytes; **c,** Reprojection error.

Malaga), the state-of-the-art LUD algorithm for global translation is not satisfactory, while the LiGT algorithm does not appear to be affected and recovers quality 3D scenes that are very close to those of LUD-BA or LiGT-PA.

*Local small translation.* Special cases exist in the Lund dataset where the camera rotates in a fixed location to take multiple photos (see, for example, #36: Sri-Veeramakaliamman and #37: Statue-of-Liberty in Fig. 3). These local small translation motions commonly lead to unsatisfactory results for the LUD algorithm, but they are handled well by the LiGT algorithm.

*Time and memory statistics.* The running time excludes data file reading and writing. The memory cost is calculated using the Intel VTune Profiler [48].

## 5. Conclusion

This study presents a pose-only representation for the multiple-view imaging geometry and discovers that it is linearly related to camera translation by the LiGT constraint. The proposed LiGT algorithm not only produces the global translation efficiently and accurately but, together with the PA algorithm, can further enhance the accuracy and robustness (for example, to critical camera motions) of recovering the camera pose and 3D scene structure. This work is believed to significantly reduce the efficiency and robustness challenges encountered in 3D vision computation. In applications where global rotations can be provided accurately, nonlinear optimization processes may not be required for camera poses and the 3D scene structure. Consequently, the computational cost would be mitigated by several orders of magnitude, hopefully opening a door to future lightweight 3D visual computation on personal devices or microchips.

# SUPPLEMENTARY INFORMATION

# A Pose-only Solution to Visual Reconstruction and Navigation

Qi Cai[1#], Lilian Zhang[2#], Yuanxin Wu[1#*], Wenxian Yu[1], Dewen Hu[2*]

[1] Shanghai Key Laboratory of Navigation and Location-based Services, School of Electronic Information and Electrical Engineering, Shanghai Jiao Tong University, China.

[2] College of Intelligent Science and Technology, National University of Defense Technology, China.

#These authors contributed equally to this work.

*Corresponding authors (yuanxin.wu@sjtu.edu.cn; dwhu@nudt.edu.cn).

# Contents

1  **FIG. 1 | REPRESENTATIVE RESULTS OF THE LUND/OPENSLAM DATASETS.**

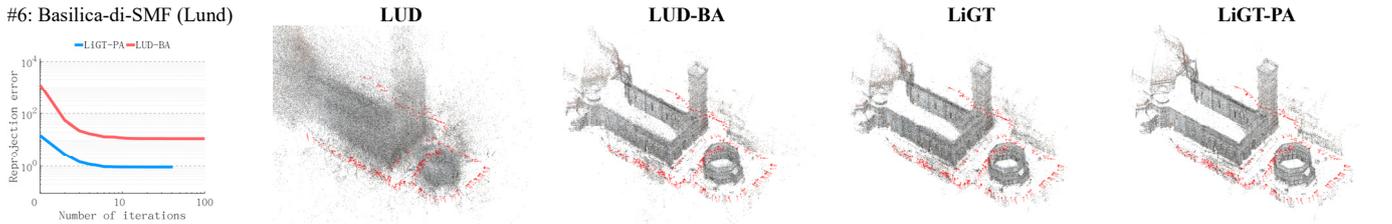
#6: Basilica-di-SMF (Lund) — LUD — LUD-BA — LiGT — LiGT-PA

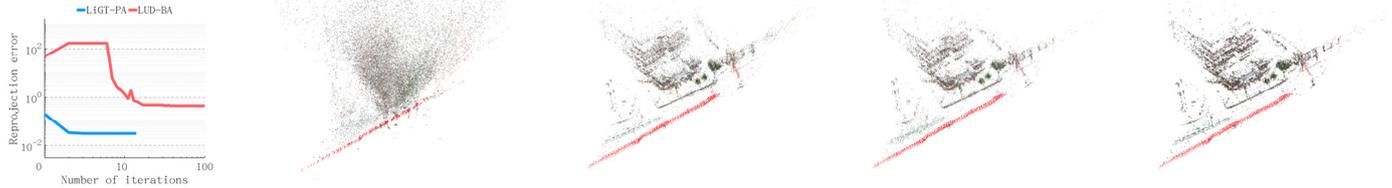
#8: Buddha-temple (Lund)

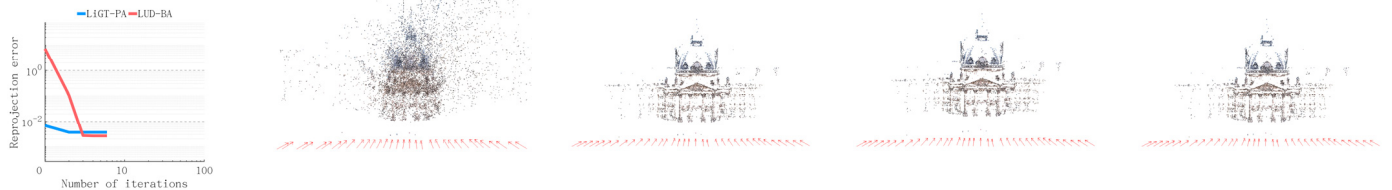
#9: De-Guerre (Lund)

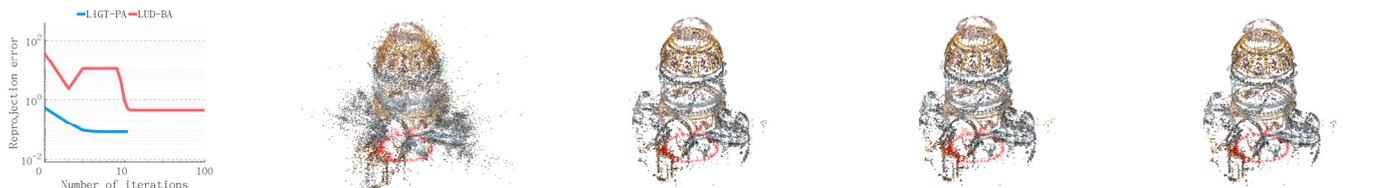
#13: Eglise-interior (Lund)

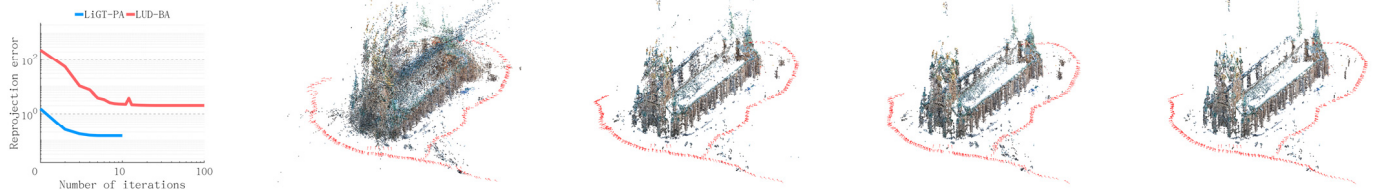
#24: Linkoping-Cathedral (Lund)

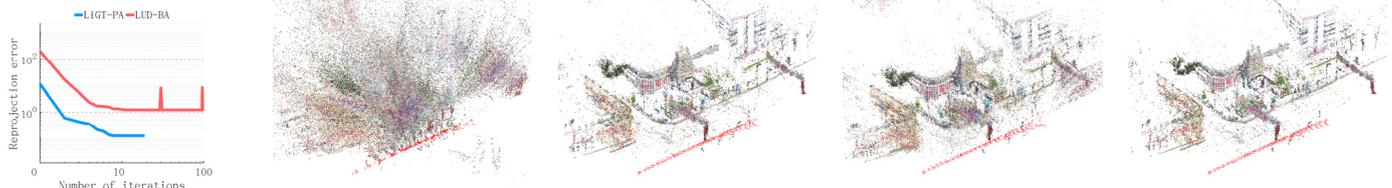
#36: Sri-Veeramakaliamman (Lund)

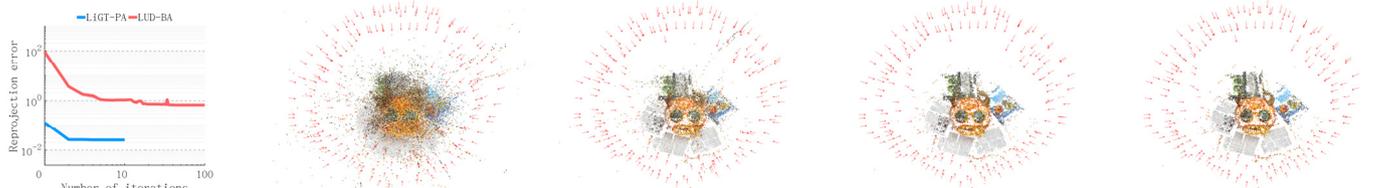
#38: The-Pumpkin (Lund)

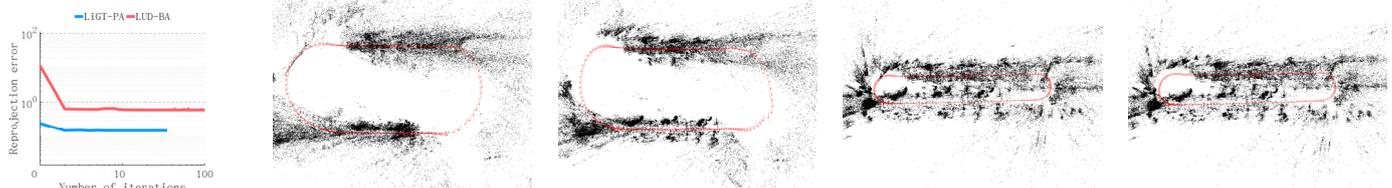
#50: Malaga (OpenSLAM)

The charts in the first column show the reprojection errors for LiGT-PA and LUD-BA as a function of the number of iterations performed during the optimisation process. Charts in the remaining four columns show the recovered 3D scenes by LUD, LUD-BA, LiGT, and LiGT-PA, respectively. The 3D scenes were recovered analytically in LiGT and LiGT-PA, and by traditional triangulation in LUD. The LiGT algorithm enables faster convergence and smaller reprojection errors for PA, and is very close to BA and PA in 3D scene quality. Note that there are collinear motions in #8: Buddha-temple, #9: De-Guerre, and #36: Sri-Veeramakaliamman, local linear motion in #50: Malaga, and local small translation in #36: Sri-Veeramakaliamman.

# FIG. 2 | OTHER RESULTS OF THE LUND DATASET.

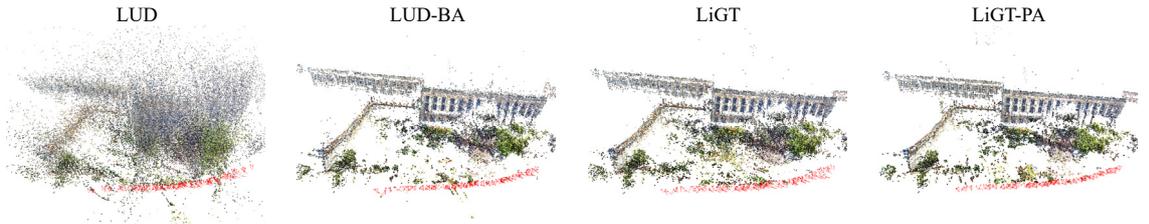

#1: West-Side-gardens
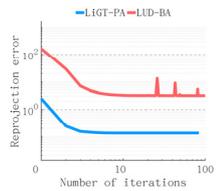

#2: Alcatraz-courtyard
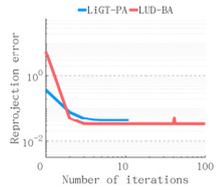
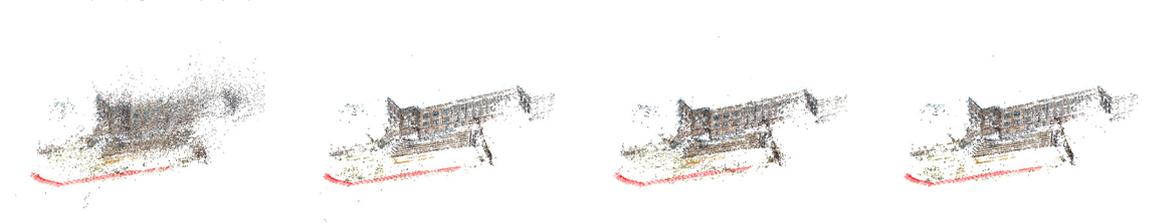

#3: Alcatraz-water-tower
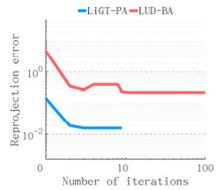
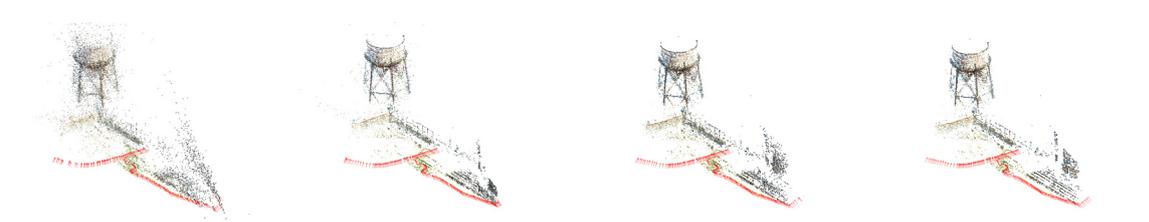

#4: Barcelona
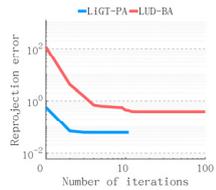
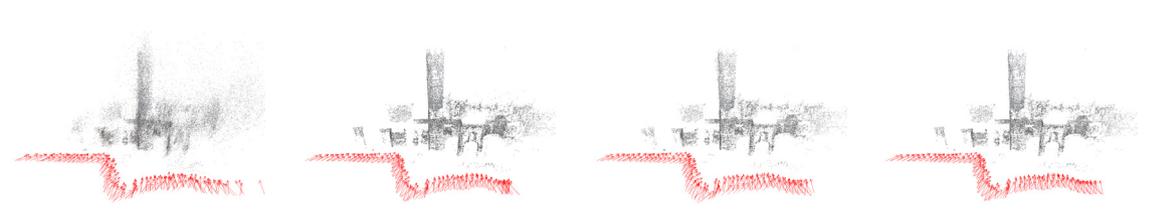

#5: Basilica-di-San-Petronio
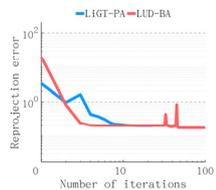
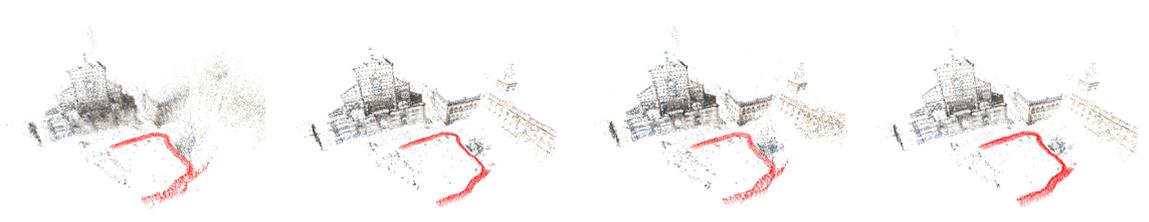

#7: Buddha
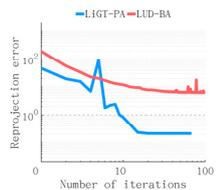
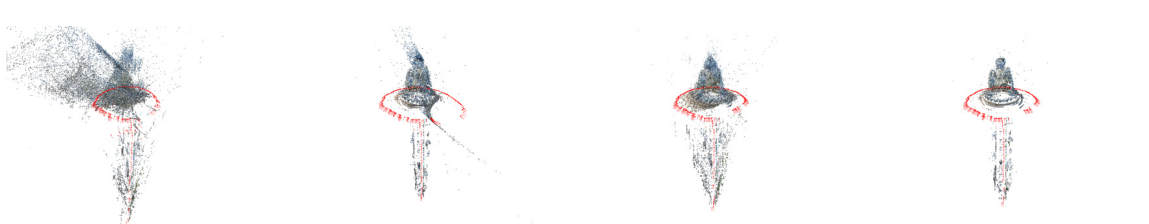

#10: Doge's-Palace
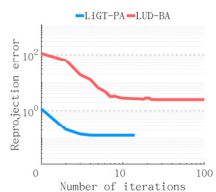
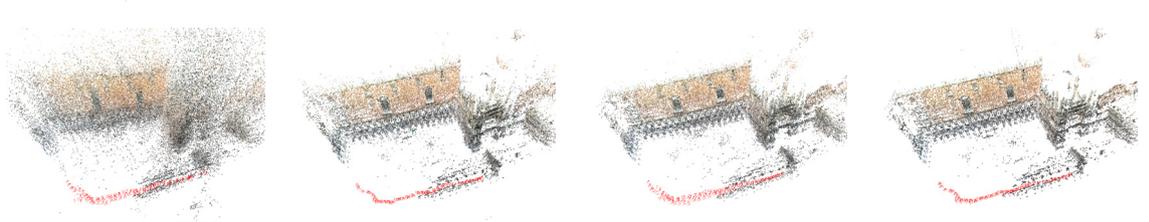

#11: Door
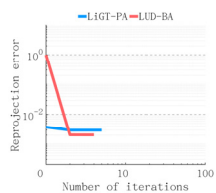
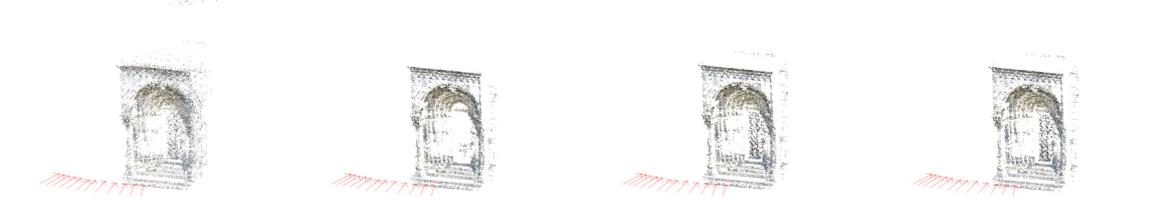

#12: Eglise 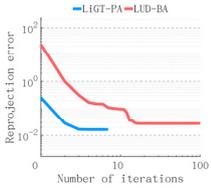 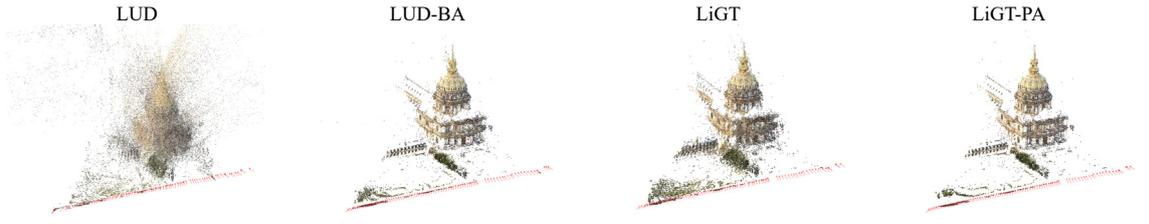

#14: Filbyter 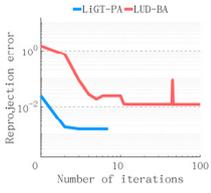 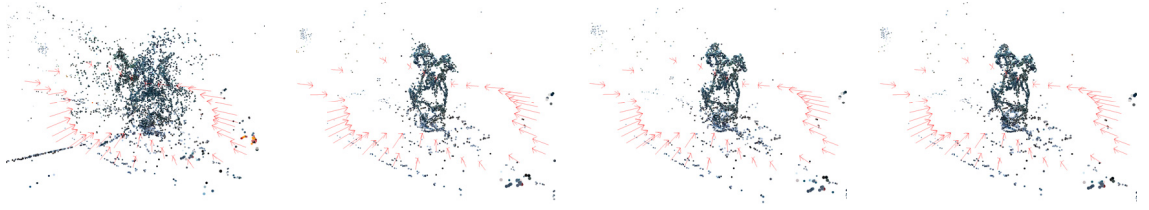

#15: Fine-Arts 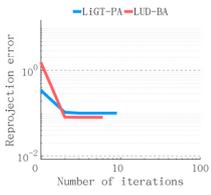 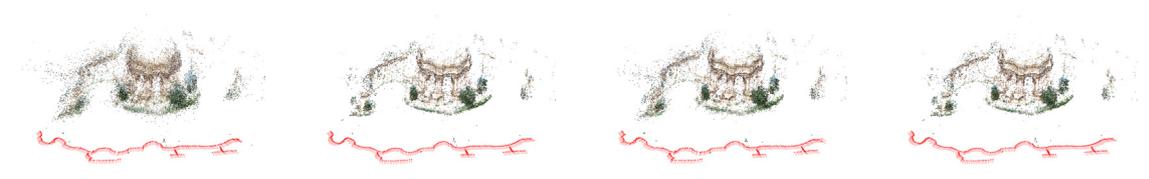

#16: Fort-Channing-gate 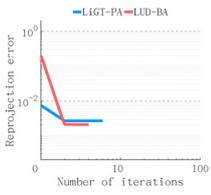 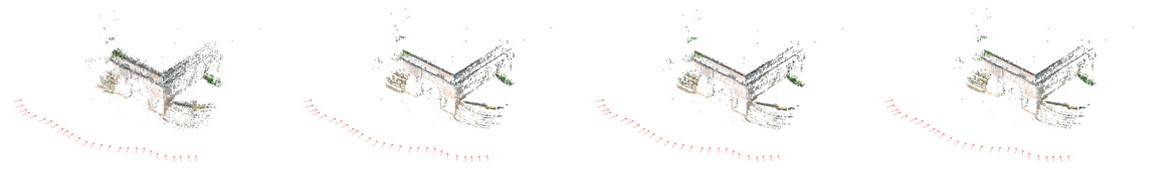

#17: Golden-statue 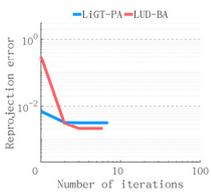 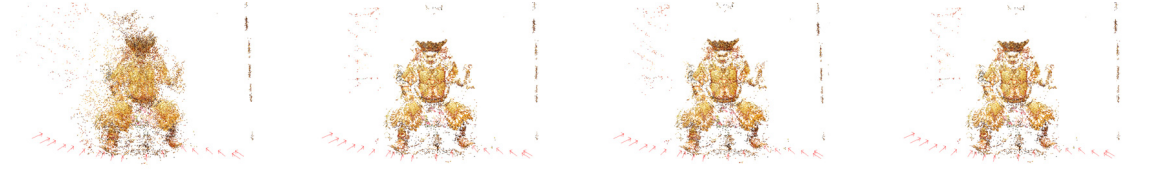

#18: Goteborg 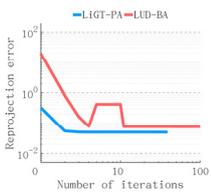 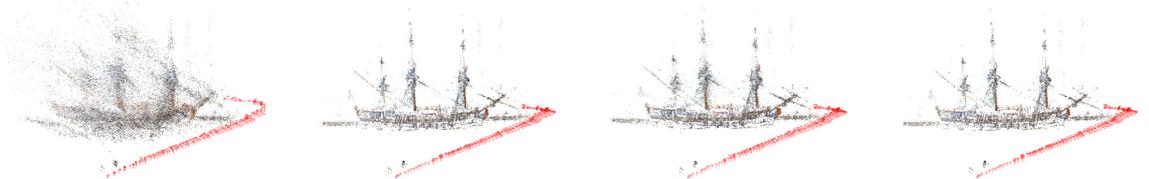

#19: GustavIIAdolf 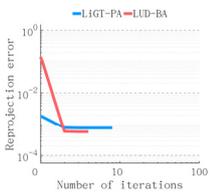 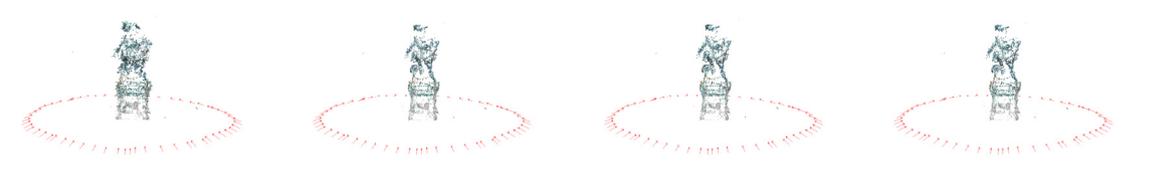

#21: Kronan 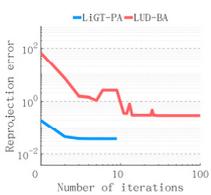 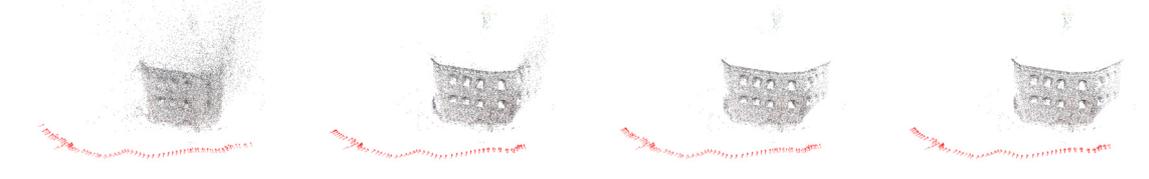

|  | LUD | LUD-BA | LiGT | LiGT-PA |
|---|---|---|---|---|
| #22: Lejonet 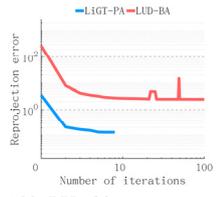 | 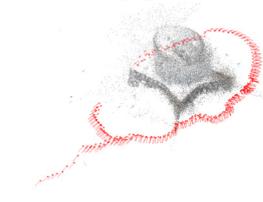 | 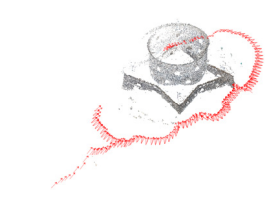 | 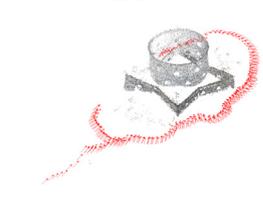 | 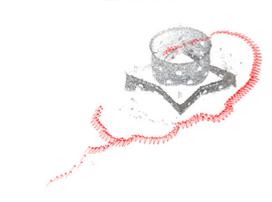 |
| #23: LUsphinx 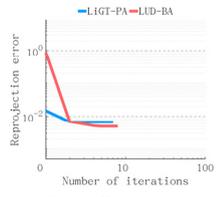 | 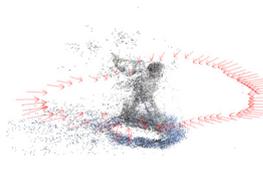 | 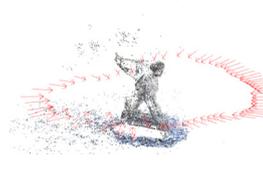 | 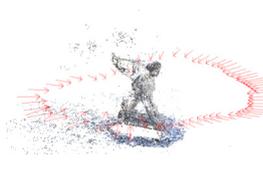 | 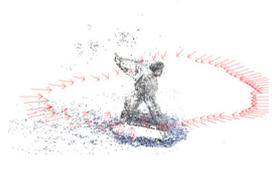 |
| #25: Lund-Cathedral 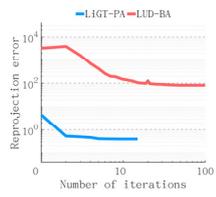 | 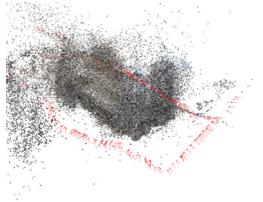 | 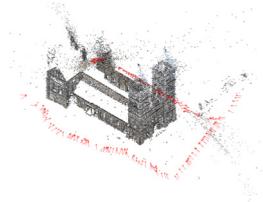 | 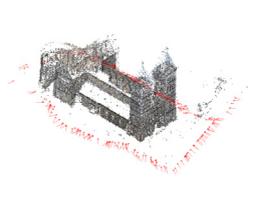 | 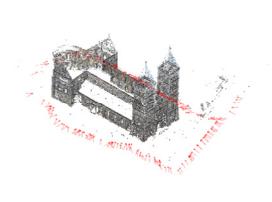 |
| #26: Nijo 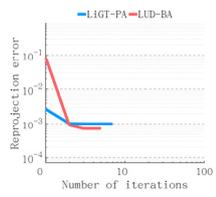 | 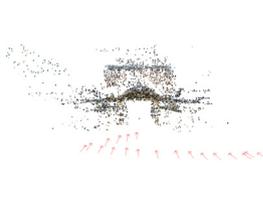 | 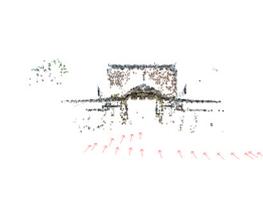 | 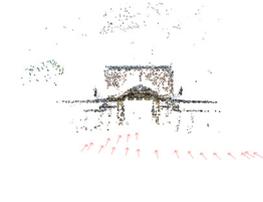 | 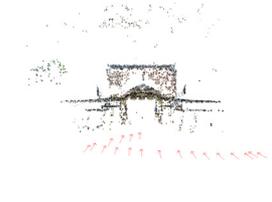 |
| #27: Nikolai 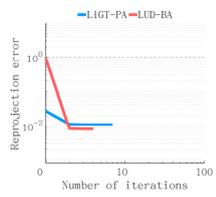 | 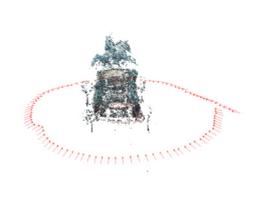 | 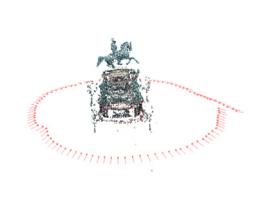 | 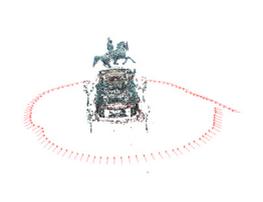 | 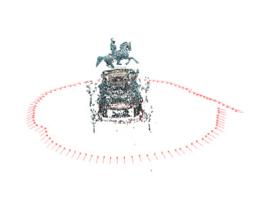 |
| #28: Orebro 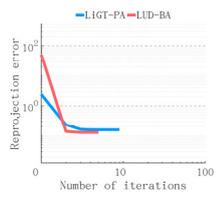 | 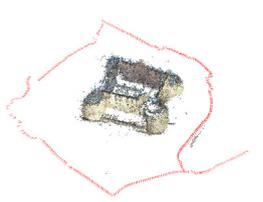 | 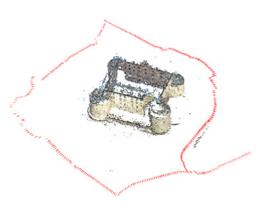 | 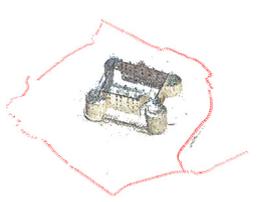 | 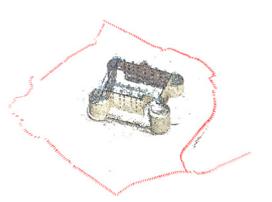 |
| #29: Park-gate 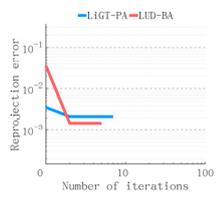 | 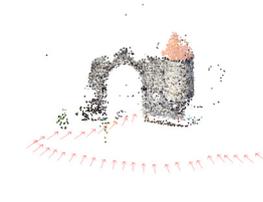 | 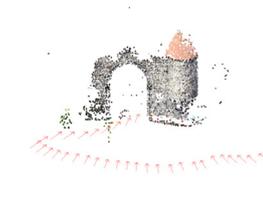 | 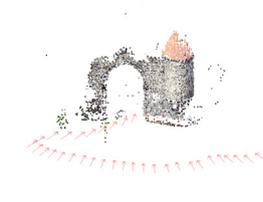 | 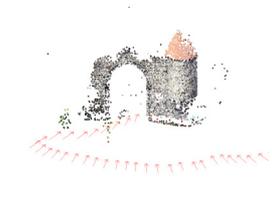 |
| #30: Plaza-de-Armas 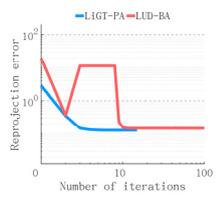 | 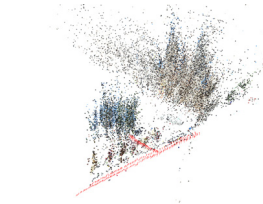 | 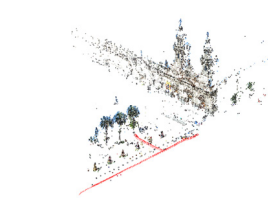 | 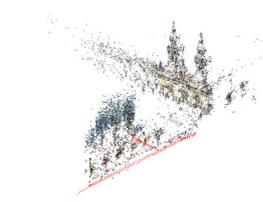 | 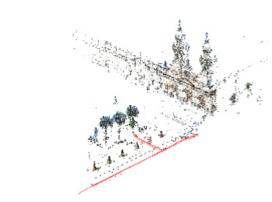 |

| | LUD | LUD-BA | LiGT | LiGT-PA |
|---|---|---|---|---|
| #31: Porta-San-Donato 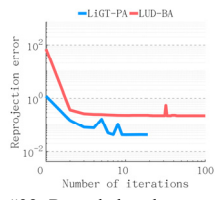 | 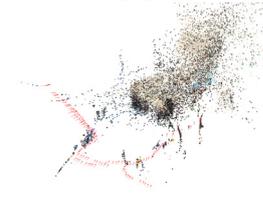 | 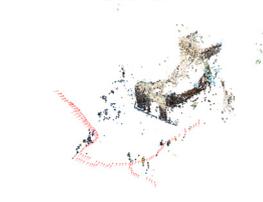 | 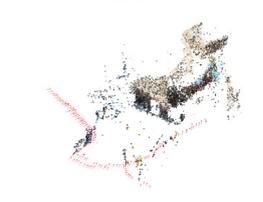 | 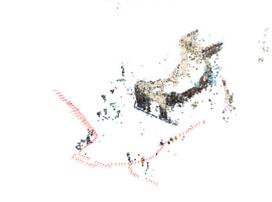 |
| #32: Round church 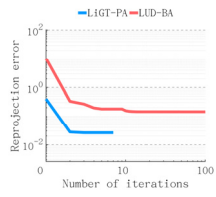 | 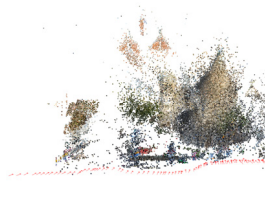 | 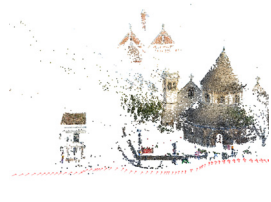 | 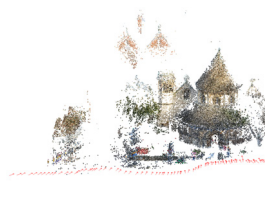 | 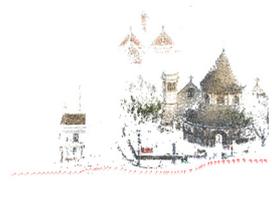 |
| #33: Smolny 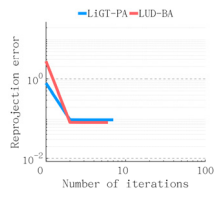 | 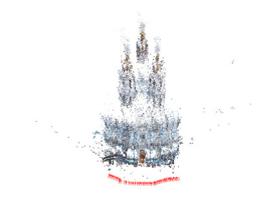 | 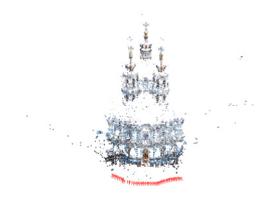 | 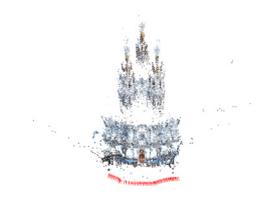 | 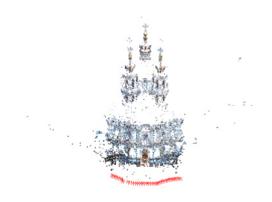 |
| #34: Sri_Mariamman 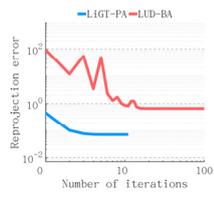 | 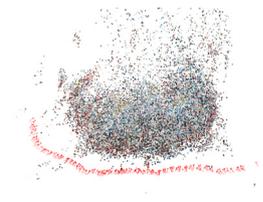 | 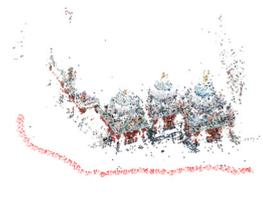 | 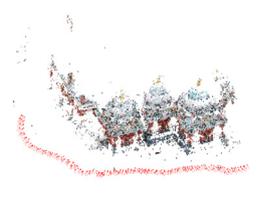 | 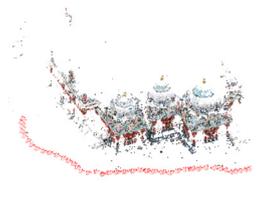 |
| #35: Sri-Thendayuthapani 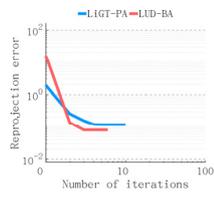 | 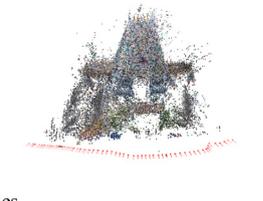 | 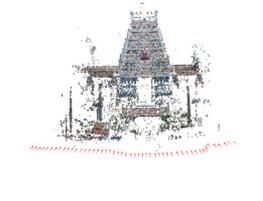 | 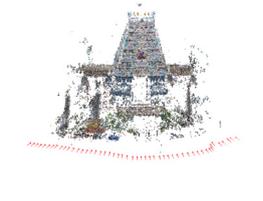 | 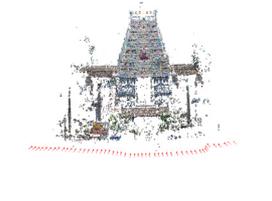 |
| #39: Thian-Hook-Keng-temples 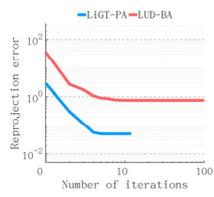 | 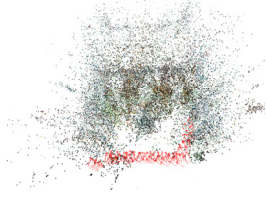 | 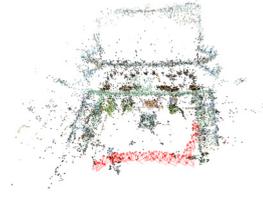 | 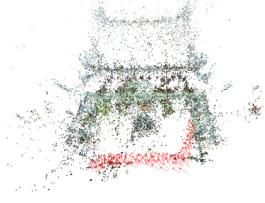 | 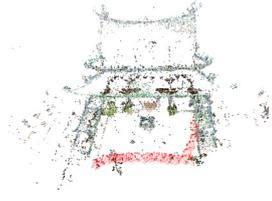 |
| #41: University-of-Toronto 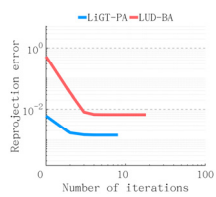 | 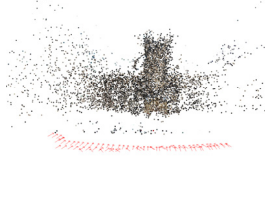 | 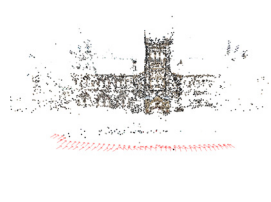 | 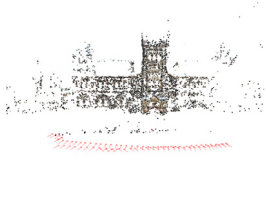 | 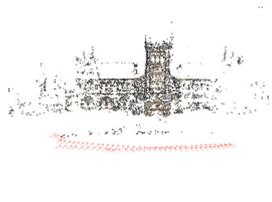 |
| #42: UrbanII 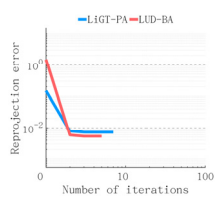 | 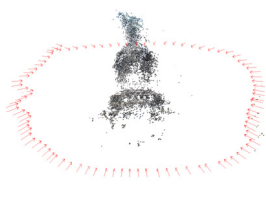 | 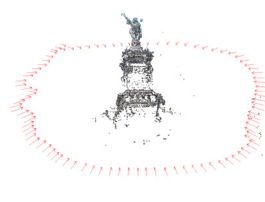 | 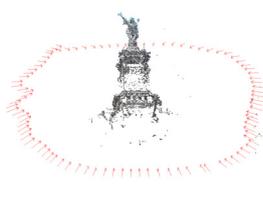 | 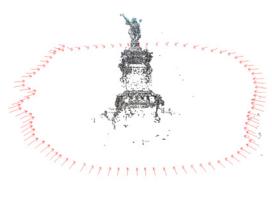 |

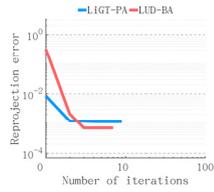 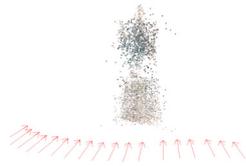 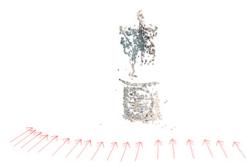 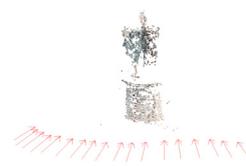 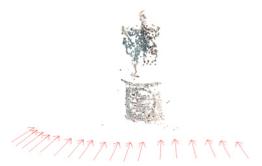
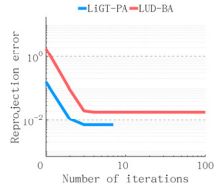 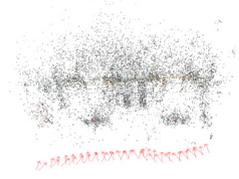 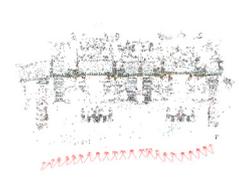 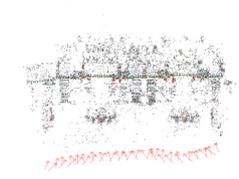 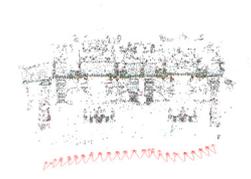

The charts in the first column show the reprojection errors for LiGT-PA and LUD-BA as a function of the number of iterations performed during the optimisation process. Charts in the remaining four columns show the recovered 3D scenes by LUD, LUD-BA, LiGT, and LiGT-PA, respectively. The 3D scenes were recovered analytically in LiGT and LiGT-PA, and by traditional triangulation in LUD.

1 **FIG. 3 | OTHER RESULTS OF THE OPENSLAM DATASET.**

| | LUD | LUD-BA | LiGT | LiGT-PA |
|---|---|---|---|---|
| #46: College 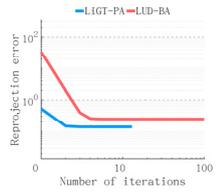 | 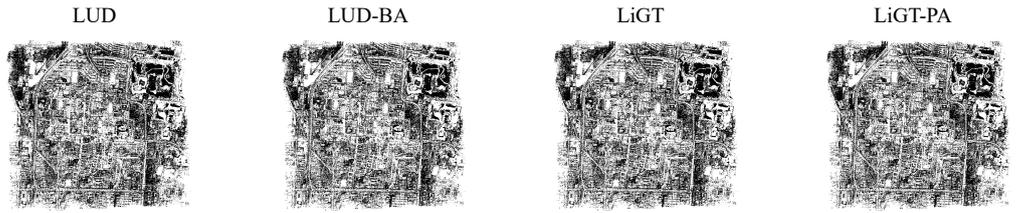 | | | |
| #47: Dunhuan 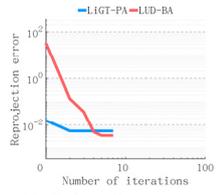 | 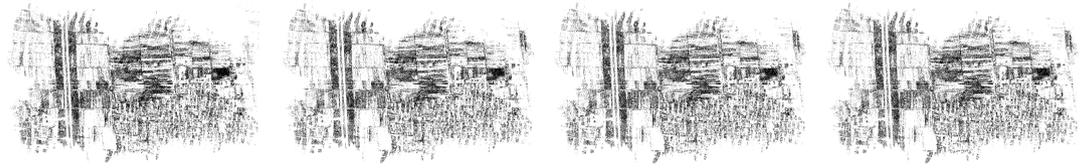 | | | |
| #48: Fake-pile 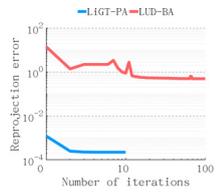 | 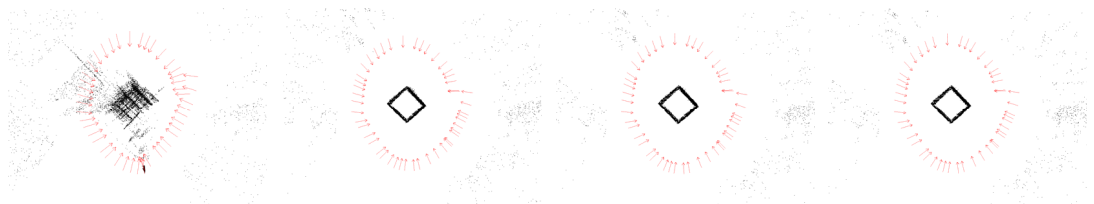 | | | |
| #49: Jinan 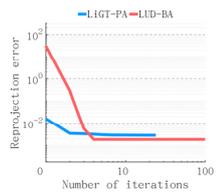 | 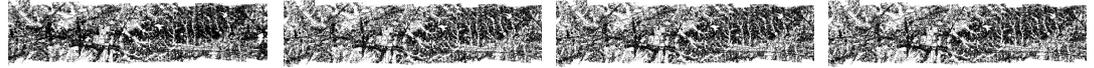 | | | |
| #51: Toronto 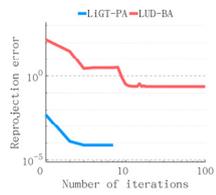 | 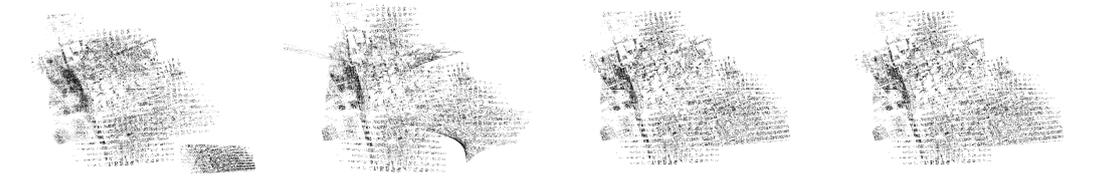 | | | |
| #52: Usyd_main_quad 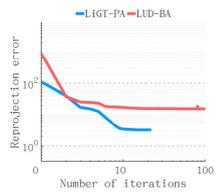 | 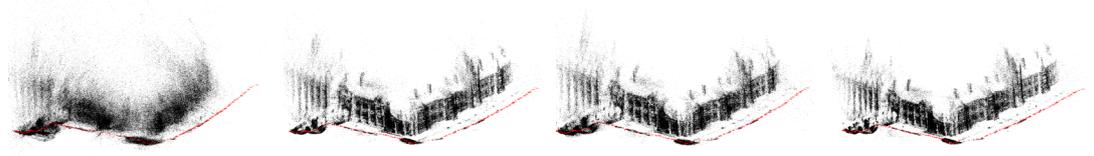 | | | |
| #53: Vaihingen 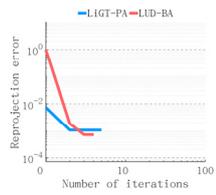 | 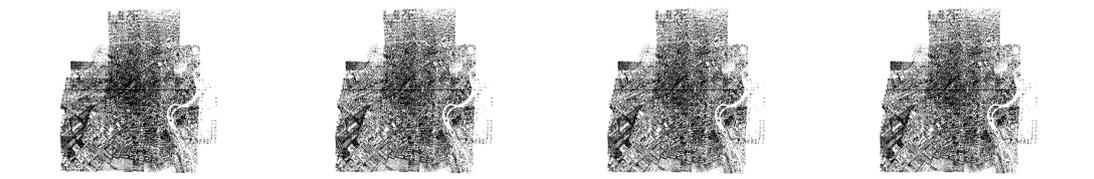 | | | |
| #54: Victoria_cottage | | | | |

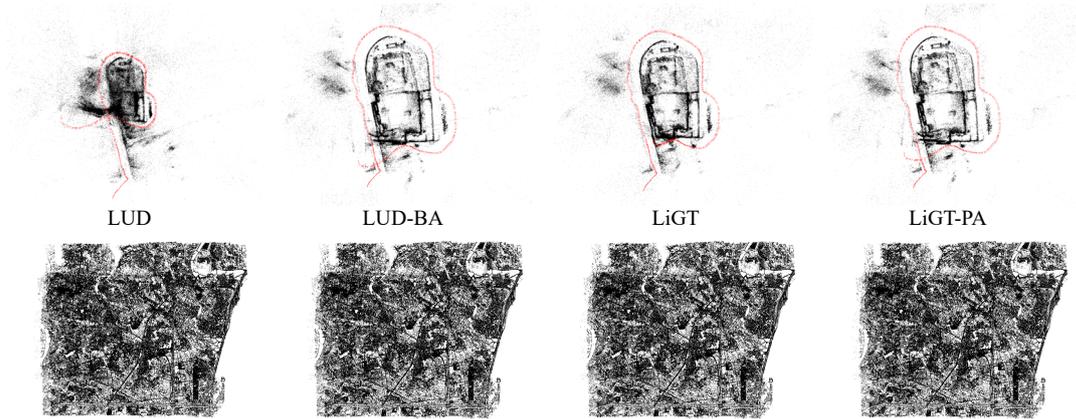

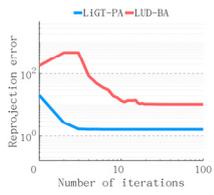

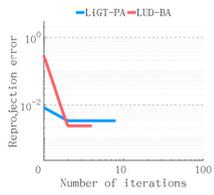

The charts in the first column show the reprojection errors for LiGT-PA and LUD-BA as a function of the number of iterations performed during the optimisation process. Charts in the remaining four columns show the recovered 3D scenes by LUD, LUD-BA, LiGT, and LiGT-PA, respectively. The 3D scenes were recovered analytically in LiGT and LiGT-PA, and by traditional triangulation in LUD.

1   TABLE 1 | TIME COST

| # | data (Lund) | cams | pts | obs | time (LiGT) | time (LiGT-PA) | iteration (LiGT-PA) | time (LUD-BA) | iteration (LUD-BA) | time ratio (LiGT-PA/LiGT) | time ratio (LUD-BA/LiGT) |
|---|---|---|---|---|---|---|---|---|---|---|---|
| 1 | Alcatraz-West-Side-gardens | 419 | 65072 | 697968 | 0.418 | 86.285 | 84 | 234.287 | 101 | 206.5 | 560.8 |
| 2 | Alcatraz-courtyard | 133 | 23674 | 321554 | 0.046 | 3.880 | 11 | 83.217 | 101 | 84.1 | 1803.1 |
| 3 | Alcatraz-water-tower | 172 | 14828 | 169618 | 0.065 | 1.314 | 9 | 47.975 | 101 | 20.3 | 742.5 |
| 4 | Barcelona | 177 | 30367 | 401584 | 0.058 | 3.281 | 11 | 134.467 | 101 | 57.0 | 2334.5 |
| 5 | Basilica-di-San-Petronio | 334 | 46035 | 806486 | 0.218 | 16.597 | 23 | 238.457 | 101 | 76.3 | 1096.4 |
| 6 | Basilica-di-SMF | 1774 | 564904 | 4851293 | 4.944 | 608.280 | 41 | 2445.910 | 101 | 123.0 | 494.7 |
| 7 | Buddha | 322 | 156356 | 920284 | 0.113 | 134.123 | 68 | 483.883 | 101 | 1184.9 | 4275.0 |
| 8 | Buddha-temple | 162 | 27920 | 201150 | 0.050 | 2.206 | 14 | 91.387 | 101 | 44.1 | 1828.8 |
| 9 | De-Guerre | 35 | 13477 | 106440 | 0.003 | 0.465 | 6 | 3.245 | 6 | 145.8 | 1016.6 |
| 10 | Doge's-Palace | 241 | 67107 | 820330 | 0.122 | 7.332 | 14 | 374.276 | 101 | 60.2 | 3070.6 |
| 11 | Door | 12 | 17650 | 140585 | 0.000 | 0.434 | 5 | 2.516 | 4 | 986.6 | 5723.6 |
| 12 | Eglise | 85 | 84792 | 619743 | 0.024 | 2.119 | 7 | 192.993 | 101 | 87.6 | 7972.9 |
| 13 | Eglise-interior | 496 | 29314 | 407967 | 0.339 | 4.814 | 11 | 190.215 | 101 | 14.2 | 561.3 |
| 14 | Filbyter | 40 | 21150 | 53028 | 0.002 | 0.915 | 7 | 16.131 | 101 | 471.1 | 8305.3 |
| 15 | Fine-Arts | 281 | 30723 | 550028 | 0.221 | 3.560 | 9 | 18.614 | 6 | 16.1 | 84.2 |
| 16 | Fort-Channing-gate | 27 | 23627 | 149430 | 0.002 | 0.590 | 6 | 2.915 | 4 | 367.7 | 1816.2 |
| 17 | Golden-statue | 18 | 39989 | 123059 | 0.001 | 0.540 | 7 | 3.721 | 6 | 521.1 | 3591.9 |
| 18 | Goteborg | 179 | 25655 | 298204 | 0.059 | 21.786 | 39 | 99.479 | 101 | 368.0 | 1680.3 |
| 19 | GustavIIAdolf | 57 | 5813 | 39015 | 0.004 | 0.724 | 8 | 0.818 | 4 | 176.1 | 198.9 |
| 20 | King's-College | 328 | 238449 | 3034113 | 0.197 | 104.843 | 51 | 1343.690 | 101 | 533.2 | 6833.2 |
| 21 | Kronan | 131 | 28371 | 430408 | 0.048 | 2.358 | 9 | 142.600 | 101 | 49.0 | 2964.5 |
| 22 | Lejonet | 368 | 74423 | 934344 | 0.180 | 7.405 | 8 | 290.651 | 101 | 41.1 | 1612.4 |
| 23 | LUsphinx | 70 | 32668 | 148069 | 0.008 | 0.850 | 7 | 6.016 | 8 | 103.1 | 730.2 |
| 24 | Linkoping-Cathedral | 538 | 202737 | 1810690 | 0.227 | 12.291 | 10 | 686.701 | 101 | 54.2 | 3026.3 |
| 25 | Lund-Cathedral | 1206 | 159055 | 2322955 | 5.021 | 40.305 | 15 | 1540.220 | 101 | 8.0 | 306.7 |
| 26 | Nijo | 19 | 7348 | 31123 | 0.001 | 0.224 | 7 | 0.833 | 5 | 216.1 | 804.3 |
| 27 | Nikolai | 98 | 37857 | 332239 | 0.015 | 1.580 | 7 | 6.467 | 4 | 108.8 | 445.6 |
| 28 | Orebro | 761 | 53857 | 1322851 | 1.455 | 11.391 | 9 | 51.795 | 5 | 7.8 | 35.6 |
| 29 | Park-gate | 34 | 9099 | 65889 | 0.003 | 0.383 | 7 | 1.423 | 5 | 140.8 | 523.6 |
| 30 | Plaza-de-Armas | 240 | 26969 | 519049 | 0.143 | 5.365 | 15 | 215.965 | 101 | 37.6 | 1513.7 |
| 31 | Porta-San-Donato | 141 | 25490 | 265347 | 0.049 | 7.032 | 19 | 79.139 | 101 | 142.8 | 1606.7 |
| 32 | Round-church | 92 | 84643 | 629033 | 0.029 | 2.213 | 7 | 141.120 | 101 | 76.2 | 4856.0 |
| 33 | Smolny | 131 | 51115 | 827085 | 0.039 | 3.093 | 7 | 24.593 | 6 | 79.2 | 629.7 |
| 34 | Sri-Mariamman | 222 | 56220 | 467301 | 0.078 | 4.223 | 11 | 140.022 | 101 | 54.3 | 1799.8 |
| 35 | Sri-Thendayuthapani | 98 | 88849 | 1010965 | 0.026 | 5.050 | 10 | 31.216 | 6 | 191.7 | 1184.6 |
| 36 | Sri-Veeramakaliamman | 157 | 130013 | 951427 | 0.052 | 8.806 | 19 | 227.872 | 101 | 170.2 | 4404.9 |

| # | data | cams | pts | obs | time (LiGT) | time (LiGT-PA) | iteration (LiGT-PA) | time (LUD-BA) | iteration (LUD-BA) | time ratio (LiGT-PA/LiGT) | time ratio (LUD-BA/LiGT) |
|---|---|---|---|---|---|---|---|---|---|---|---|
| 37 | Statue-of-Liberty | 133 | 49248 | 203917 | 0.016 | 46.854 | 50 | 68.954 | 101 | 2877.5 | 4234.7 |
| 38 | The-Pumpkin | 196 | 69341 | 267379 | 0.065 | 3.706 | 10 | 98.843 | 101 | 57.1 | 1523.2 |
| 39 | Thian-Hook-Keng-temple | 138 | 34288 | 228181 | 0.031 | 2.053 | 12 | 70.475 | 101 | 65.5 | 2248.9 |
| 40 | UWO | 691 | 97326 | 1324691 | 0.998 | 11.158 | 10 | 761.110 | 101 | 11.2 | 762.7 |
| 41 | University-of-Toronto | 77 | 7087 | 44562 | 0.006 | 0.841 | 8 | 5.073 | 18 | 143.9 | 867.8 |
| 42 | UrbanII | 96 | 22284 | 183784 | 0.013 | 1.138 | 7 | 4.258 | 5 | 89.4 | 334.4 |
| 43 | Vasa | 18 | 4249 | 15544 | 0.001 | 0.452 | 9 | 0.689 | 7 | 483.4 | 736.8 |
| 44 | Ystad-Monestary | 290 | 139951 | 1514025 | 0.122 | 7.916 | 9 | 570.254 | 101 | 64.9 | 4673.1 |
| 45 | Yueh-Hai-Ching-Temple | 43 | 13774 | 76314 | 0.004 | 0.426 | 7 | 20.339 | 101 | 97.5 | 4653.9 |
| # | data (OpenSLAM) | cams | pts | obs | time (LiGT) | time (LiGT-PA) | iteration (LiGT-PA) | time (LUD-BA) | iteration (LUD-BA) | time ratio (LiGT-PA/LiGT) | time ratio (LUD-BA/LiGT) |
| 46 | College | 468 | 1236502 | 3107524 | 0.070 | 42.023 | 13 | 978.974 | 101 | 599.9 | 13975.4 |
| 47 | DunHuan | 63 | 250782 | 597289 | 0.004 | 2.425 | 7 | 29.736 | 7 | 599.4 | 7348.4 |
| 48 | Fake-pile | 47 | 11318 | 26050 | 0.002 | 0.587 | 10 | 23.323 | 101 | 247.9 | 9846.2 |
| 49 | Jinan | 76 | 1228959 | 2864740 | 0.005 | 22.444 | 24 | 764.939 | 101 | 4390.8 | 149647.7 |
| 50 | Malaga | 170 | 58404 | 167285 | 0.014 | 79.590 | 35 | 97.678 | 101 | 5869.9 | 7204.0 |
| 51 | Toronto | 13 | 113685 | 239279 | 0.001 | 0.409 | 7 | 95.316 | 101 | 536.3 | 125115.5 |
| 52 | Usyd-main-quad | 424 | 227615 | 1607082 | 0.210 | 28.654 | 22 | 479.589 | 101 | 136.7 | 2287.9 |
| 53 | Vaihingen | 20 | 554169 | 1201982 | 0.001 | 1.584 | 5 | 31.824 | 4 | 1425.7 | 28633.9 |
| 54 | Victoria-cottage | 400 | 153632 | 890057 | 0.102 | 205.554 | 100 | 393.529 | 101 | 2023.6 | 3874.1 |
| 55 | Village | 90 | 1849740 | 4877796 | 0.007 | 14.391 | 8 | 130.252 | 4 | 2114.4 | 19137.3 |

(cams: cameras; pts: 3D points; obs: image point observations.)

Table 1 lists the running time (in seconds) of LiGT, LiGT-PA, and LUD-BA, respectively, and the iteration times of LiGT-PA and LUD-BA in the Lund and OpenSLAM tests. The running time ratios of LiGT-PA and LUD-BA with respect to LiGT are calculated in the two right-most columns and are plotted in Fig. 2.

1 TABLE 2 | REPROJECTION ERROR

| # | Data (Lund) | cams | pts | obs | initial reprojection error | | end reprojection error | | | |
|---|---|---|---|---|---|---|---|---|---|---|
| | | | | | LiGT-BA&PA | LUD-BA&PA | LiGT-BA | LUD-BA | LiGT-PA | LUD-PA |
| 1 | Alcatraz-West-Side-gardens | 419 | 65072 | 697968 | **2.478** | 162.351 | **0.103** | 3.180 | 0.134 | 0.127 |
| 2 | Alcatraz-courtyard | 133 | 23674 | 321554 | **0.379** | 6.032 | 0.035 | 0.033 | 0.044 | 0.044 |
| 3 | Alcatraz-water-tower | 172 | 14828 | 169618 | **0.147** | 4.539 | **0.012** | 0.225 | 0.016 | 0.016 |
| 4 | Barcelona | 177 | 30367 | 401584 | **0.579** | 118.330 | **0.047** | 0.388 | 0.060 | 0.060 |
| 5 | Basilica-di-San-Petronio | 334 | 46035 | 806486 | **3.382** | 20.403 | **0.156** | 0.173 | 1.130 | 0.198 |
| 6 | Basilica-di-SMF | 1774 | 564904 | 4851293 | **15.214** | 6844.949 | **0.789** | 11.563 | 0.911 | 0.857 |
| 7 | Buddha | 322 | 156356 | 920284 | 46.615 | 188.014 | **0.408** | 6.306 | 3.701 | 2.505 |
| 8 | Buddha-temple | 162 | 27920 | 201150 | **0.209** | 46.621 | **0.024** | 0.454 | 0.031 | 0.031 |
| 9 | De-Guerre | 35 | 13477 | 106440 | **0.007** | 8.086 | 0.003 | 0.003 | 0.004 | 0.004 |
| 10 | Doge's-Palace | 241 | 67107 | 820330 | **1.191** | 111.959 | **0.112** | 2.518 | 0.129 | 0.263 |
| 11 | Door | 12 | 17650 | 140585 | **0.004** | 1.081 | 0.002 | 0.002 | 0.003 | 0.003 |
| 12 | Eglise | 85 | 84792 | 619743 | **0.259** | 27.713 | **0.011** | 0.028 | 0.016 | 0.016 |
| 13 | Eglise-interior | 496 | 29314 | 407967 | **0.541** | 39.318 | **0.070** | 0.450 | 0.080 | 0.081 |
| 14 | Filbyter | 40 | 21150 | 53028 | **0.025** | 1.498 | **0.001** | 0.012 | 0.002 | 0.002 |
| 15 | Fine-Arts | 281 | 30723 | 550028 | **0.353** | 1.636 | 0.084 | 0.084 | 0.104 | 0.105 |
| 16 | Fort-Channing-gate | 27 | 23627 | 149430 | **0.008** | 0.209 | 0.002 | 0.002 | 0.003 | 0.003 |
| 17 | Golden-statue | 18 | 39989 | 123059 | **0.007** | 0.349 | 0.002 | 0.002 | 0.003 | 0.003 |
| 18 | Goteborg | 179 | 25655 | 298204 | **0.329** | 18.969 | **0.041** | 0.077 | 0.050 | 0.051 |
| 19 | GustavIIAdolf | 57 | 5813 | 39015 | **0.002** | 0.159 | 0.001 | 0.001 | 0.001 | 0.001 |
| 20 | King's-College | 328 | 238449 | 3034113 | **13.880** | 2493.362 | **0.396** | 18.496 | **0.343** | 94.881 |
| 21 | Kronan | 131 | 28371 | 430408 | **0.194** | 72.943 | **0.031** | 0.293 | 0.037 | 0.037 |
| 22 | Lejonet | 368 | 74423 | 934344 | **3.582** | 270.539 | **0.110** | 2.467 | **0.132** | 3.316 |
| 23 | LUsphinx | 70 | 32668 | 148069 | **0.015** | 0.923 | 0.005 | 0.005 | 0.007 | 0.007 |
| 24 | Linkoping-Cathedral | 538 | 202737 | 1810690 | **1.547** | 232.937 | **0.125** | 2.051 | 0.147 | 0.197 |
| 25 | Lund-Cathedral | 1206 | 159055 | 2322955 | **4.586** | 3294.693 | **0.311** | 97.377 | **0.399** | 67131.299 |
| 26 | Nijo | 19 | 7348 | 31123 | **0.003** | 0.095 | 0.001 | 0.001 | 0.001 | 0.001 |
| 27 | Nikolai | 98 | 37857 | 332239 | **0.027** | 1.000 | 0.008 | 0.008 | 0.011 | 0.011 |
| 28 | Orebro | 761 | 53857 | 1322851 | **2.458** | 53.101 | 0.128 | 0.128 | 0.157 | 0.157 |
| 29 | Park-gate | 34 | 9099 | 65889 | **0.004** | 0.038 | 0.001 | 0.001 | 0.002 | 0.002 |
| 30 | Plaza-de-Armas | 240 | 26969 | 519049 | 2.828 | 20.451 | **0.103** | 0.144 | 0.126 | 0.126 |
| 31 | Porta-San-Donato | 141 | 25490 | 265347 | **1.220** | 70.170 | **0.031** | 0.229 | 0.049 | 0.039 |
| 32 | Round-church | 92 | 84643 | 629033 | **0.402** | 10.276 | **0.019** | 0.145 | 0.026 | 0.026 |
| 33 | Smolny | 131 | 51115 | 827085 | 0.780 | 2.819 | 0.084 | 0.084 | 0.098 | 0.098 |
| 34 | Sri-Mariamman | 222 | 56220 | 467301 | **0.455** | 94.694 | **0.056** | 0.657 | 0.070 | 0.070 |
| 35 | Sri-Thendayuthapani | 98 | 88849 | 1010965 | 1.916 | 16.567 | 0.081 | 0.081 | 0.118 | 0.119 |
| 36 | Sri-Veeramakaliamman | 157 | 130013 | 951427 | **11.596** | 201.514 | **0.154** | 1.227 | 0.127 | 0.255 |
| 37 | Statue-of-Liberty | 133 | 49248 | 203917 | **1.393** | 172.478 | **0.014** | 0.747 | 0.017 | 0.035 |
| 38 | The-Pumpkin | 196 | 69341 | 267379 | **0.128** | 92.353 | **0.020** | 0.661 | 0.025 | 0.025 |

| # | data | cams | pts | obs | initial reprojection error | | end reprojection error | | | |
|---|---|---|---|---|---|---|---|---|---|---|
| | | | | | LiGT-BA&PA | LUD-BA&PA | LiGT-BA | LUD-BA | LiGT-PA | LUD-PA |
| 39 | Thian-Hook-Keng-temple | 138 | 34288 | 228181 | **2.936** | 35.184 | 0.129 | 0.758 | 0.050 | 0.493 |
| 40 | UWO | 691 | 97326 | 1324691 | **3.058** | 819.236 | **0.132** | 28.172 | **0.158** | 3.152 |
| 41 | University-of-Toronto | 77 | 7087 | 44562 | **0.006** | 0.566 | 0.001 | 0.007 | 0.001 | 0.001 |
| 42 | UrbanII | 96 | 22284 | 183784 | **0.149** | 1.569 | 0.005 | 0.005 | 0.007 | 0.007 |
| 43 | Vasa | 18 | 4249 | 15544 | **0.009** | 0.360 | 0.001 | 0.001 | 0.001 | 0.001 |
| 44 | Ystad-Monastery | 290 | 139951 | 1514025 | **1.951** | 474.913 | **0.162** | 1.704 | 0.191 | 0.190 |
| 45 | Yueh-Hai-Ching-Temple | 43 | 13774 | 76314 | **0.160** | 1.802 | 0.010 | 0.019 | 0.007 | 0.007 |
| # | data (OpenSLAM) | cams | pts | obs | initial reprojection error | | end reprojection error | | | |
| | | | | | LiGT-BA&PA | LUD-BA&PA | LiGT-BA | LUD-BA | LiGT-PA | LUD-PA |
| 46 | College | 468 | 1236502 | 3107524 | 0.531 | 33.728 | 0.079 | 0.246 | 0.138 | 0.138 |
| 47 | DunHuan | 63 | 250782 | 597289 | **0.013** | 48.234 | 0.003 | 0.003 | 0.005 | 0.005 |
| 48 | Fake-pile | 47 | 11318 | 26050 | **0.002** | 13.747 | **0.001** | 0.492 | **0.001** | 0.010 |
| 49 | Jinan | 76 | 1228959 | 2864740 | **0.005** | 49.547 | 0.002 | 0.002 | 0.003 | 0.003 |
| 50 | Malaga | 170 | 58404 | 167285 | **0.229** | 10.404 | **0.027** | 0.606 | 0.150 | 0.227 |
| 51 | Toronto | 13 | 113685 | 239279 | **0.001** | 290.775 | 0.000 | 0.232 | 0.001 | 0.000 |
| 52 | Usyd-main-quad | 424 | 227615 | 1607082 | **107.747** | 879.497 | 2.485 | 15.537 | 3.206 | 3.217 |
| 53 | Vaihingen | 20 | 554169 | 1201982 | **0.001** | 1.133 | 0.001 | 0.001 | 0.001 | 0.001 |
| 54 | Victoria-cottage | 400 | 153632 | 890057 | 19.258 | 170.191 | **1.273** | 12.691 | 1.621 | 1.626 |
| 55 | Village | 90 | 1849740 | 4877796 | **0.005** | 0.377 | 0.002 | 0.002 | 0.003 | 0.003 |

Table 2 lists the initial and the final reprojection errors of four optimisation algorithms: LUD-BA, LUD-PA, LiGT-BA, and LiGT-PA. Note that their initial reprojection errors are exactly those of the corresponding initialisation algorithms. The reprojection errors have been regularised uniformly, for all algorithms, by way of BA's minimisation function, using their own estimates of camera poses and 3D feature coordinates. The reprojection errors are partly plotted in Fig. 2. In fact, we tested and found that all algorithms of LUD, 1DSfM, LinearSfM, and OpenMVG are generally consistent under normal scenarios, but LUD performs the best under abnormal scenarios. Therefore, only LUD is compared with LiGT here. The reprojection error columns are given colour backgrounds for clear comparison and, in the same colour region, the bold-faced numbers indicate that the reprojection errors brought about by LiGT and LUD are different by over one order of magnitude.

- The reprojection error of LiGT is over one order of magnitude smaller than that of LUD in most test data, and even better than those of LUD-BA/PA in #20: King's-College, #25: Lund-Cathedral, and #48: Fake-pile.
- For all test data, the reprojection errors of LiGT-BA/PA are consistently superior to those of LUD-BA/PA.

TABLE 3 | SCATTERING DEGREE OF RECONSTRUCTED 3D POINTS

| # | Data (Lund) | LiGT | LiGT-BA | LiGT-PA | LUD | LUD-BA | LUD-PA |
|---|---|---|---|---|---|---|---|
| 1 | Alcatraz-West-Side-gardens | 0.430 | 3.391 | 0.335 | 1.998 | 96.705 | 0.530 |
| 2 | Alcatraz-courtyard | 0.242 | 4.780 | 0.179 | 0.494 | 1.238 | 0.172 |
| 3 | Alcatraz-water-tower | 0.343 | 0.547 | 0.272 | 0.737 | 104.230 | 0.272 |
| 4 | Barcelona | 0.082 | 0.063 | 0.063 | 0.244 | 22.787 | 0.063 |
| 5 | Basilica-di-San-Petronio | 0.182 | 3.085 | 0.127 | 0.436 | 6.590 | 0.127 |
| 6 | Basilica-di-SMF | 1.084 | 4.641 | 14.976 | 99.594 | 36.779 | 18.901 |
| 7 | Buddha | 0.001 | 0.460 | 0.001 | 0.003 | 6.882 | 0.001 |
| 8 | Buddha-temple | 0.210 | 0.165 | 0.166 | 0.626 | 136.659 | 0.166 |
| 9 | De-Guerre | 0.058 | 0.055 | 0.055 | 0.209 | 0.055 | 0.055 |
| 10 | Doge's-Palace | 0.428 | 2.275 | 0.337 | 4.448 | 4018.160 | 1.189 |
| 11 | Door | 0.030 | 0.030 | 0.030 | 0.072 | 0.030 | 0.030 |
| 12 | Eglise | 0.124 | 0.078 | 0.078 | 0.342 | 1.904 | 0.078 |
| 13 | Eglise-interior | 0.704 | 0.558 | 0.556 | 1.997 | 62.421 | 0.556 |
| 14 | Filbyter | 0.221 | 0.160 | 0.160 | 2.359 | 31.924 | 0.160 |
| 15 | Fine-arts | 0.369 | 0.283 | 0.285 | 0.526 | 0.283 | 0.285 |
| 16 | Fort-Channing-gate | 0.020 | 0.018 | 0.018 | 0.042 | 0.018 | 0.018 |
| 17 | Golden-statue | 0.057 | 0.044 | 0.044 | 0.110 | 0.044 | 0.044 |
| 18 | Goteborg | 0.124 | 0.096 | 0.096 | 0.342 | 13.864 | 0.096 |
| 19 | GustavIIAdolf | 0.037 | 0.036 | 0.036 | 0.059 | 0.036 | 0.036 |
| 20 | King's-College | 0.132 | 4.272 | 0.084 | 5.292 | 339.921 | 0.203 |
| 21 | Kronan | 0.215 | 0.179 | 0.181 | 0.790 | 118.725 | 0.181 |
| 22 | Lejonet | 0.120 | 0.082 | 0.083 | 0.341 | 32.606 | 0.095 |
| 23 | LUsphinx | 0.066 | 0.058 | 0.058 | 0.116 | 0.058 | 0.058 |
| 24 | Linkoping-Cathedral | 0.037 | 0.027 | 0.027 | 0.111 | 8.586 | 0.027 |
| 25 | Lund-Cathedral | 0.048 | 0.034 | 0.034 | 0.163 | 15.314 | 0.055 |
| 26 | Nijo | 0.073 | 0.066 | 0.066 | 0.124 | 0.066 | 0.066 |
| 27 | Nikolai | 0.038 | 0.035 | 0.035 | 0.072 | 0.035 | 0.035 |
| 28 | Orebro | 0.210 | 0.171 | 0.171 | 0.396 | 0.171 | 0.171 |
| 29 | Park-gate | 0.050 | 0.048 | 0.048 | 0.069 | 0.048 | 0.048 |
| 30 | Plaza-de-Armas | 0.189 | 1.752 | 0.116 | 0.361 | 15.202 | 0.116 |
| 31 | Porta-San-Donato | 0.102 | 0.059 | 0.059 | 0.205 | 5.844 | 0.059 |
| 32 | Round-church | 0.047 | 0.029 | 0.029 | 0.147 | 10.633 | 0.029 |
| 33 | Smolny | 0.240 | 0.179 | 0.180 | 0.338 | 0.179 | 0.180 |
| 34 | Sri-Mariamman | 0.152 | 0.116 | 0.118 | 0.399 | 111.891 | 0.118 |
| 35 | Sri-Thendayuthapani | 0.441 | 0.348 | 0.355 | 0.776 | 0.348 | 0.355 |
| 36 | Sri-Veeramakaliamman | 0.364 | 2.162 | 0.238 | 0.879 | 39.661 | 0.249 |
| 37 | Statue-of-Liberty | 0.032 | 0.408 | 0.020 | 0.078 | 21.858 | 0.021 |
| 38 | The-Pumpkin | 0.012 | 0.009 | 0.010 | 0.048 | 14.942 | 0.010 |
| 39 | Thian-Hook-Keng-temple | 0.161 | 16.660 | 0.089 | 0.439 | 561.589 | 0.102 |
| 40 | UWO | 0.201 | 0.143 | 0.144 | 8.108 | 244.222 | 0.207 |
| 41 | University-of-Toronto | 0.138 | 0.112 | 0.110 | 0.415 | 0.172 | 0.110 |
| 42 | UrbanII | 0.107 | 0.087 | 0.088 | 0.140 | 0.087 | 0.088 |
| 43 | Vasa | 0.047 | 0.041 | 0.041 | 0.097 | 0.041 | 0.041 |
| 44 | Ystad-Monestary | 0.209 | 0.139 | 0.145 | 0.813 | 60.364 | 0.145 |
| 45 | Yueh-Hai-Ching-Temple | 0.241 | 2.758 | 0.187 | 0.419 | 10.771 | 0.187 |

Table 3 lists the 3D point-scattering phenomenon for the Lund dataset, quantified by the average distance (unit: meter) of each recovered 3D feature point from its nearest neighbour. This table is plotted in Fig. 4, clockwise, in 1   ascending order of the number of image points.

2
3

# Test Results of 1DSfM Dataset

Figure 4 presents the reconstruction result of seven data and Table 4 summarizes the number of reconstructed views and tracks. Both Theia and the pose-only solution use their default parameters throughout the tests, if not explicitly stated. Their successful rates of Theia and are 3/7 (algorithm crashed in four data) and 6/7 (poor result in one data, *Ellis Island*), respectively. Specifically, they are both successful in *Montreal Notre Dame* and *Notre Dame* with comparable reconstruction quality. Note that the reprojection errors do not completely accord with the quality of reconstruction is very likely due to the remaining outliers. It should be noted that LiGT-BA performs not well in Fig. 4, as the outlier-handling pipeline of the pose-only solution keeps a large number of point observations of small $\theta_{\varsigma,\eta}$ (as shown in Table 4, the reconstructed track number by the pose-only solution is about two times that by Theia), which is problematic to the bundle adjustment taking 3D feature coordinates as optimizing parameters. While those point observations of small $\theta_{\varsigma,\eta}$ are removed, the LiGT-BA performance is significantly improved, and LiGT and LiGT-PA are less affected, as shown in Fig. 5 with *Notre Dame*. In contrast, the final BA of Theia performs quite well because the point observations of small $\theta_{\varsigma,\eta}$ are removed as well in its outlier-handling pipeline.

For all data that the pose-only solution is successful, LiGT is very close to the optimization algorithms in reconstruction quality. The reason that LiGT-BA performs unsatisfactorily throughout the test is owed to the same above-mentioned reason.

Regarding the data of *Ellis Island*, the pose-only solution performs not well by the default parameter, but can be improved when those 3D points with track length smaller than 3 are removed, as shown in Fig. 6.

FIG. 4 | REPRESENTATIVE RESULTS OF THE 1DSfM DATASET.

| Data | Opt. Iteration | LiGT | LiGT-PA | LiGT-BA | Theia |
|---|---|---|---|---|---|
| Montreal Notre Dame | 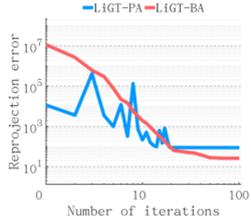 | 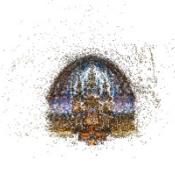 | 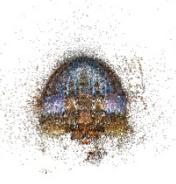 | 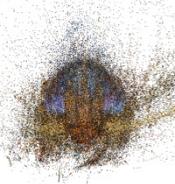 | 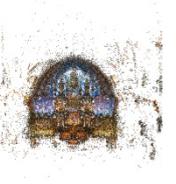 |
| Notre Dame | 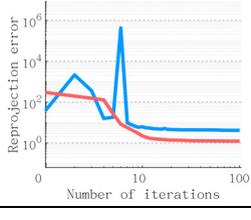 | 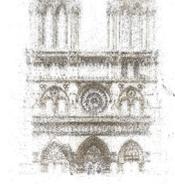 | 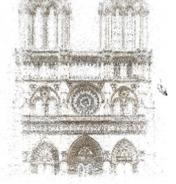 | 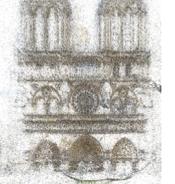 | 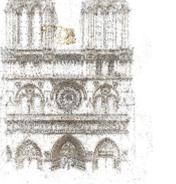 |
| Alamo | 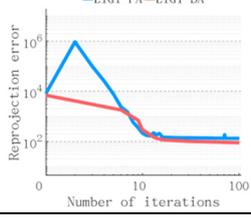 | 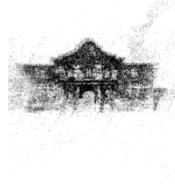 | 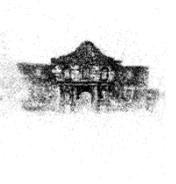 | 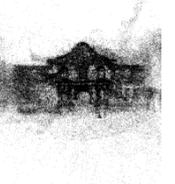 | - |
| NYC Library | 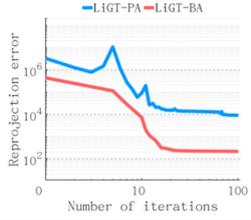 | 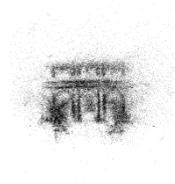 | 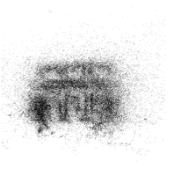 | 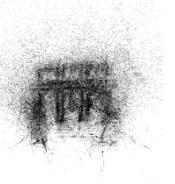 | - |
| Piazza del Popolo | 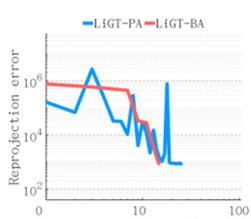 | 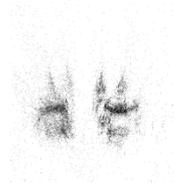 | 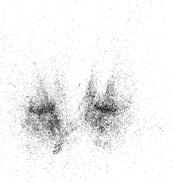 | 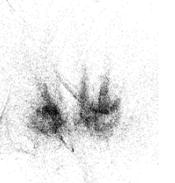 | - |
| Tower of London | 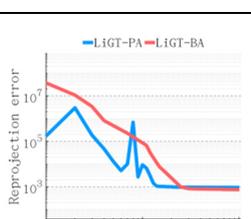 | 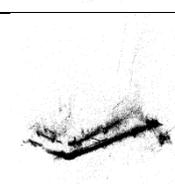 | 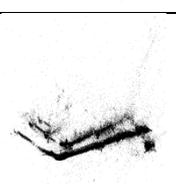 | 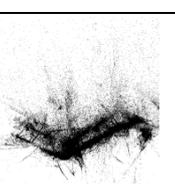 | - |
| Ellis Island | 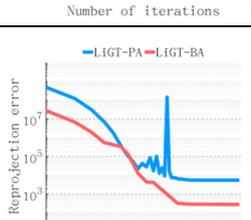 | 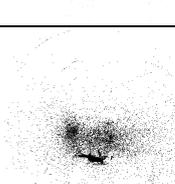 | 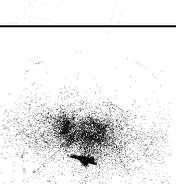 | 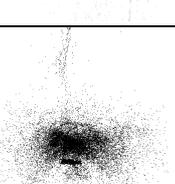 | 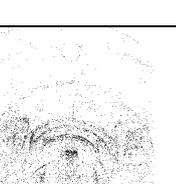 |

Both Theia and the pose-only solution use their default parameters throughout the tests. Their successful rates of Theia and are 3/7 (algorithm crashed in four data) and 6/7 (poor result in one data, *Ellis Island*), respectively. The reprojection errors are calculated by BA and PA, respectively.

FIG. 5 | NOTRE DAME RESULT BEFORE AND AFTER PARAMETER TUNING.

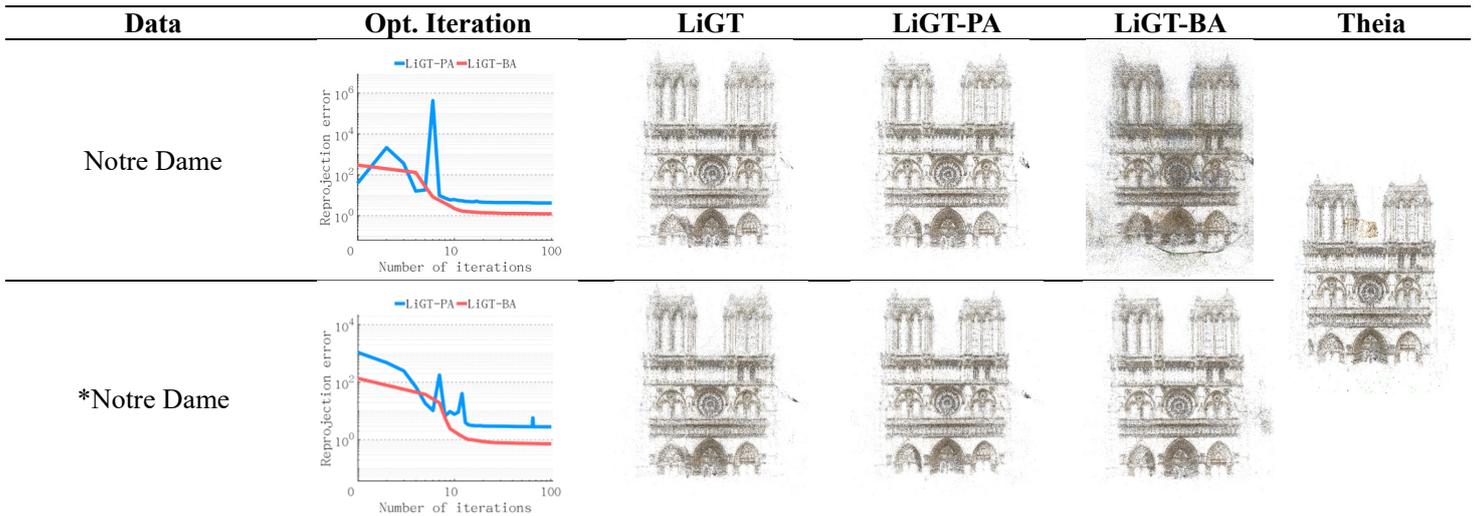

When those point observations of small $\theta_{\varsigma,\eta}$ are removed, the LiGT-BA performance is significantly improved, and LiGT and LiGT-PA are less affected.

**FIG. 6 | ELLIS ISLAND RESULT BEFORE AND AFTER PARAMETER TUNING.**

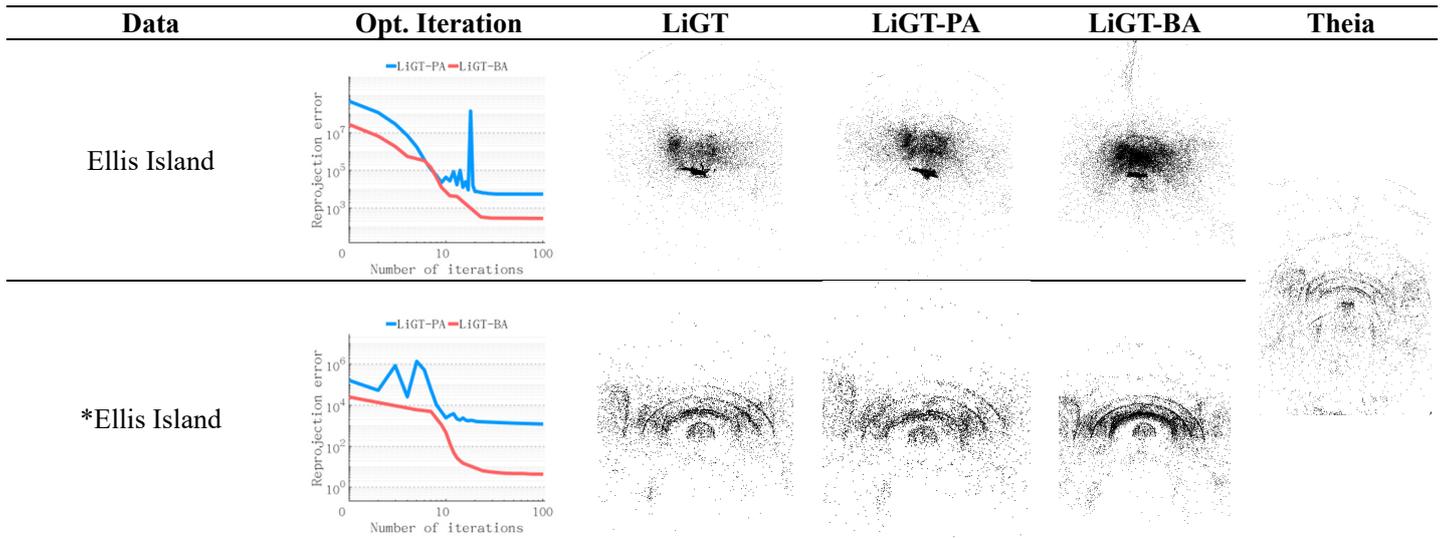

By removing 3D feature points of tracking length smaller than 3, the Ellis Island result of the pose-only solution is improved.

TABLE 4 | NUMBER OF ESTIMATED VIEWS AND TRACKS BY THEIA AND POSE-ONLY SOLUTION

| Data | Method | # Estimated views | # Input views | # Estimated tracks | # Input tracks |
|---|---|---|---|---|---|
| Montreal Notre Dame | Theia | 458 | 474 | 141181 | 337088 |
| | Pose-only | 458 | | 275223 | |
| Notre Dame | Theia | 546 | 553 | 228676 | 587692 |
| | Pose-only | 547 | | 519423 | |
| *Notre Dame | Pose-only | 546 | | 218241 | |
| Alamo | Theia | - | 627 | - | 318946 |
| | Pose-only | 584 | | 287774 | |
| NYC Library | Theia | - | 376 | - | 180176 |
| | Pose-only | 351 | | 151783 | |
| Piazza del Popolo | Theia | - | 354 | - | 98253 |
| | Pose-only | 340 | | 81370 | |
| Tower of London | Theia | - | 508 | - | 295360 |
| | Pose-only | 477 | | 267961 | |
| Ellis Island | Theia | 233 | 247 | 10746 | 108795 |
| | Pose-only | 226 | | 78189 | |
| *Ellis Island | Pose-only | 225 | | 13591 | |
| *parameter-tuned tests | | | | | |